%% file: chapters/main.tex
\documentclass[sigconf,authorversion,nonacm]{acmart}
\AtBeginDocument{%
  }

\usepackage{multirow}
\usepackage{enumitem}
\usepackage{tcolorbox}
\usepackage{xcolor}
\usepackage{subcaption}
\usepackage{soul}
\usepackage{makecell}

\newcommand{\rev}[1]{\textcolor{black}{#1}} 

\begin{document}

\newcommand{\name}{\textsf{EventCast}}
\title{\name: Hybrid Demand Forecasting in E-Commerce with LLM-Based Event Knowledge}

\author{Congcong Hu$^{1,}$\footnotemark[1], Yuang Shi$^{2,} $\footnotemark[1]$^{,}$\footnotemark[2], \\Fan Huang$^{1}$, Yang Xiang$^{1}$, Zhou Ye$^{1}$, Ming Jin$^{3}$, Shiyu Wang$^{1}$\footnotemark[3]}
\thanks{$^*$ Equal contribution with alphabetical ordering.}
\thanks{$\dag$ Work done during an internship at ByteDance.}
\thanks{$\ddagger$ Correspondence author.}
\def \authors{Congcong Hu et al.}
\affiliation{%
  \institution{$^1$ByteDance China \quad $^2$National University of Singapore \quad $^3$Griffith University}
  \country{}
}

\renewcommand{\shortauthors}{Shi et al.}


\input{chapters/0-Abstract}




\maketitle

\input{chapters/1-Introduction}

\input{chapters/2-Related}
\input{chapters/3-Methodology}
\input{chapters/4-Experiments}

\input{chapters/5-Conclusion}

\bibliographystyle{ACM-Reference-Format}
\bibliography{manuscript}

\appendix

\input{chapters/99-Appendix}

\end{document}

%% file: chapters/0-Abstract.tex
\begin{abstract}
Demand forecasting is a cornerstone of e-commerce operations, directly impacting inventory planning and fulfillment scheduling.
However, existing forecasting systems often fail during high-impact periods such as flash sales, holiday campaigns, and sudden policy interventions, where demand patterns shift abruptly and unpredictably.
In this paper, we introduce {\name}, a modular forecasting framework that integrates future event knowledge into time-series prediction. 
Unlike prior approaches that ignore future interventions or directly use large language models (LLMs) for numerical forecasting, {\name} leverages LLMs solely for event-driven reasoning.
Unstructured business data, which covers campaigns, holiday schedules, and seller incentives, from existing operational databases, is processed by an LLM that converts it into interpretable textual summaries leveraging world knowledge for cultural nuances and novel event combinations.
These summaries are fused with historical demand features within a dual-tower architecture, enabling accurate, explainable, and scalable forecasts. 
Deployed on real-world e-commerce scenarios spanning 4 countries of 160 regions over 10 months, {\name} achieves up to 86.9\% and 97.7\% improvement on MAE and MSE compared to the variant without event knowledge, and reduces MAE by up to 57.0\% and MSE by 83.3\% versus the best industrial baseline during event-driven periods.
{\name} has deployed into real-world industrial pipelines since March 2025, offering a practical solution for improving operational decision-making in dynamic e-commerce environments.
\end{abstract}

%% file: chapters/1-Introduction.tex
\begin{figure}[t!]
    \centering
    \includegraphics[width=0.97\linewidth]{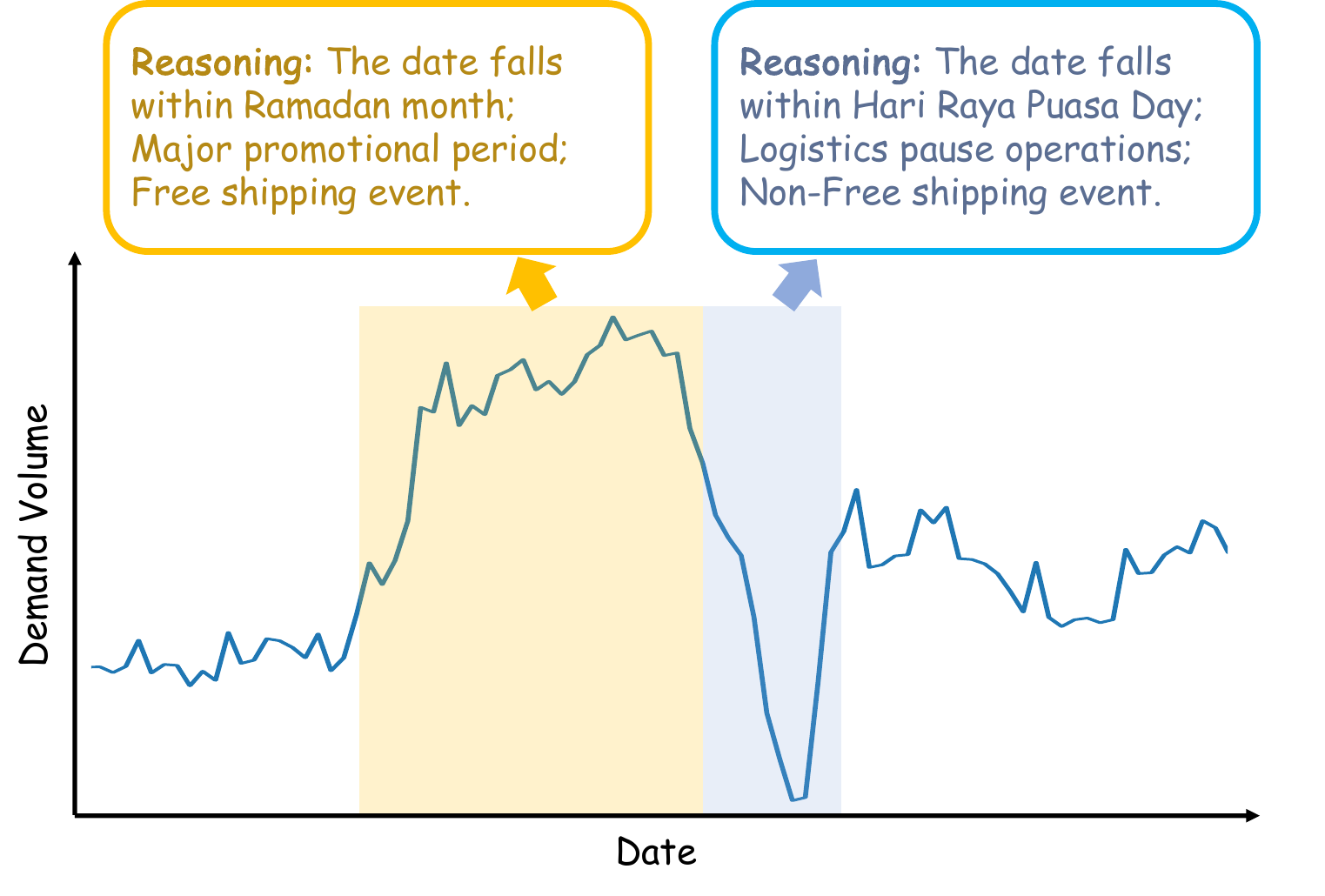}
    \vspace{-1em}
    \caption{The demand volume affected by sales promotions and religious holiday. Reasoning examples are shown. During Ramadan, demand surges sharply in the days leading up to Hari Raya Puasa as it is in a major promotional period, followed by a steep drop as people shift their focus to religious events, while businesses and logistics pause operations. }
    \label{fig:volatile_illustration}
\end{figure}

\section{Introduction}

Demand forecasting is the backbone of e-commerce supply chain management, which drives vital downstream tasks like restocking inventory, warehouse scheduling, and order fulfillment~\cite{mentzer2001defining,wang2025enhancing}. Accurate and timely forecasts balance resource efficiency with service quality, ensuring that products are available where and when they are needed~\cite{ye2022gaia,zhang2022co,ding2022multi}. However, e-commerce forecasting differs significantly from traditional retail settings, presenting unique challenges, as shown below.

\textit{Non-stationary and Event-driven Demand Patterns}. E-commerce demand is inherently non-stationary, disrupted by flash sales, coupon pushes, free-shipping promotions, and influencer campaigns, as well as external factors such as religious and public holidays.
These events concentrate revenue into short windows, creating surges or drops that cannot be predicted from historical patterns alone.
Figure~\ref{fig:volatile_illustration} illustrates how promotional campaigns and cultural holidays overlap to cause complex demand fluctuations.

\textit{Non-structural Knowledge Extraction}.
Although platforms often know future promotions, seller incentives, and holidays in advance, encoding their impact into forecasting models is non-trivial.
Planned information is often recorded in unstructured text that may contain typos, making string-matching-based feature extraction infeasible.
Conventional approaches rely on categorical flags or rule-based heuristics, which struggle with overlapping campaigns, cultural nuances, or novel promotion combinations. Forecasting systems thus often underperform during high-variance periods.

\textit{Non-interpretable Forecasting}. 
E-commerce business rules, promotional strategies, and market behaviors shift frequently.
Operational personnel need to understand why forecasts change in response to known events. Black-box models, which fail to directly provide transparent reasoning, struggle to gain trust in decision-making loops.
\rev{Interpretability is crucial for deployment. Warehouse managers may override opaque predictions based on intuition; interpretable knowledge reasoning enables rapid root cause analysis when forecasts deviate; and human-readable traces facilitate communication with teams without requiring technical expertise.}

Recent deep learning advances~\cite{benidis2022deep,nie2022time,liu2023itransformer,ansari2024chronos,woo2024unified} have improved predictive accuracy under regular conditions, but rely heavily on past demand signals under a ``self-stimulation assumption''~\cite{xu2024intervention}, overlooking known future interventions.
Recent research has attempted to address this gap by incorporating exogenous variables~\cite{wang2024timexer,arango2025chronosx}, but such approaches typically encode interventions as structured numerical or categorical variables, which remain overly rigid for complex real-world scenarios. 
More recently, inspired by the multi-modal strengths of LLMs, some studies~\cite{zhang2024large} have applied LLMs as forecasters. However, empirical results~\cite{tan2024language} consistently show LLMs underperform due to poor numerical precision, weak temporal consistency, and prohibitive computational costs. 
These limitations hinder the production deployment of LLMs in fast-moving e-commerce environments.

\begin{figure}[t!]
    \centering
    \includegraphics[width=0.99\linewidth]{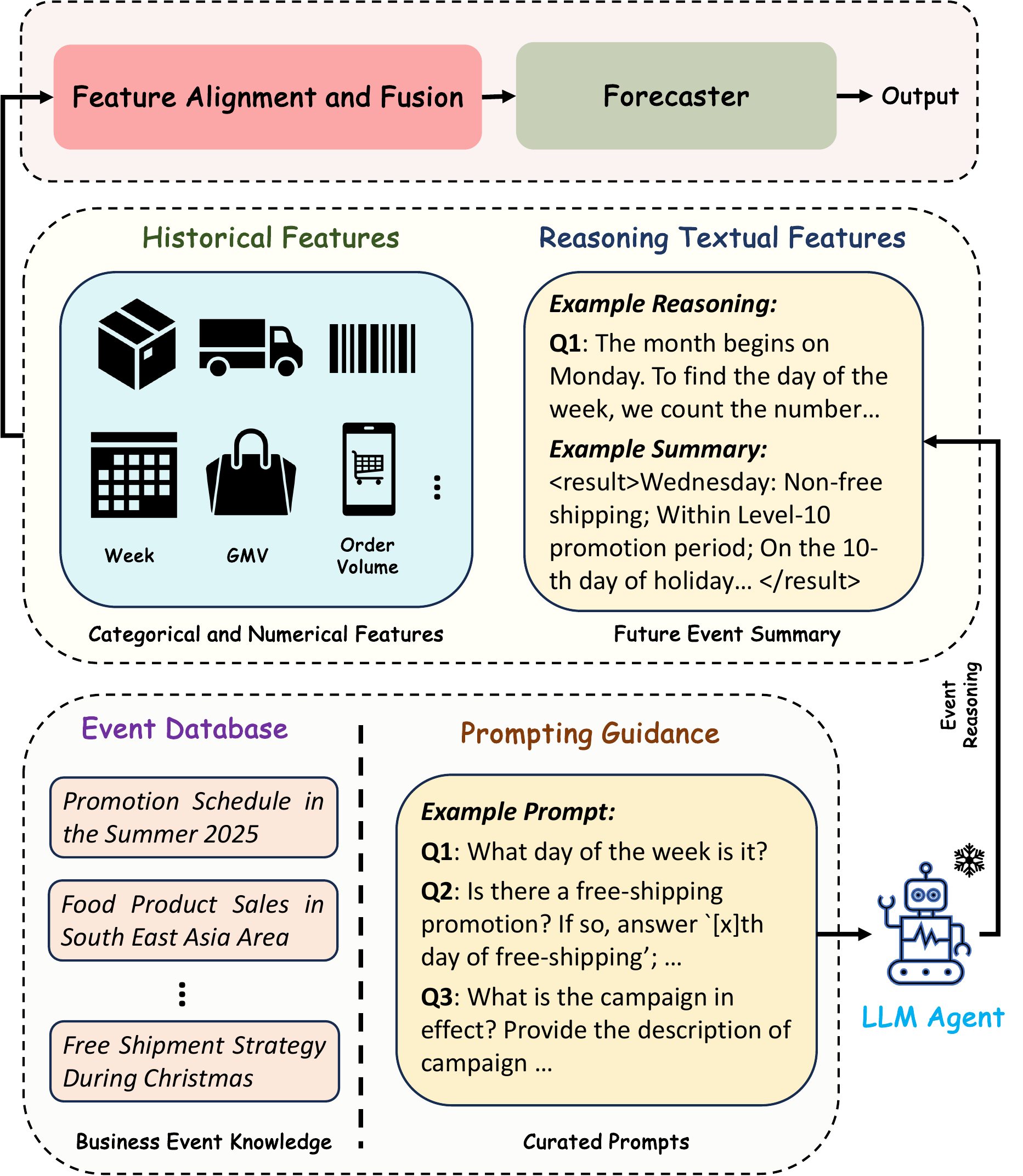}
    \caption{Pipeline overview. (1) An Event Expert Database stores business data with curated prompts. (2) An LLM Agent then acts as a domain-aware semantic reasoning agent, interpreting the business operations and converting them into interpretable textual summaries. (3) Summaries are fused with historical signals. (4) The Forecasting Model combines these embedded features for predictions.}
    \label{fig:architecture}
\end{figure}

To address these challenges, we argue that an effective forecasting solution for industrial use should be simple, easy to maintain, and efficient. Meanwhile, they also need to incorporate domain insights like promotions and holiday effects that directly affect demand.
In this paper, we propose an efficient yet effective forecasting framework {\name} that keeps forecasting straightforward and scalable, yet benefits from rich semantic reasoning powered by LLMs. 
Unlike LLM-based methods such as Time-LLM~\cite{jin2023time}, Calf~\cite{liu2025calf}, or OneFitSAll~\cite{zhou2023one}, our system does not rely on LLM embeddings or fine-tuning. Instead, the LLM is used solely for reasoning about future events, making the system lighter and easier to update. 
Specifically, \rev{we leverage an existing operational database maintained by business teams for campaign management, logistics planning, and seller coordination. E-commerce platforms inherently require such records to execute promotions and manage operations.} This expert database contains unstructured textual data describing campaigns, holiday schedules, and seller live streams. The LLM leverages this expert database to reason about each forecast target date, producing interpretable summaries of what is important about that date. 
These summaries are converted into feature embeddings and aligned with historical features in a shared feature space. A lightweight forecasting module then learns from both past trends and future event signals. Figure \ref{fig:architecture} illustrates the overall pipeline.

Our contributions are summarized as follows:
\begin{itemize}[leftmargin=*]
    \item We propose {\name}, a forecasting architecture that integrate i) historical multivariate time-series features with ii) unstructured future multivariate features distilled into textual summaries via LLM event knowledge reasoning. 
    The dual-tower design ensures future event signals remain explicit rather than diluted in the high-dimensional historical latent representation, while preserving interpretability through human-readable summaries.
    \item {\name} avoids heavy LLM components like fine-tuned LLM embeddings. Instead, it leverages only the text outputs of the LLM. This enables rapid adaptation through prompt engineering and LLM-based knowledge reasoning, and reduces reliance on manual feature engineering. Each component is independently replaceable (e.g., swap in a better LLM or Agent) for ease of deployment and maintenance at scale.
    \item We deploy {\name} across 4 countries and 160 regions over 10 months, achieving average reductions of 57.0\% in MAE and 83.3\% in MSE during event-driven periods. \rev{The framework has been running in production since March 2025, demonstrating its maintainability for industrial use.}
\end{itemize}

%% file: chapters/2-Related.tex
\section{Related Works}

\rev{
\textbf{Traditional Knowledge Incorporation}. 
Traditional methods for incorporating domain-specific knowledge into time-series forecasting require structured data inputs; for instance, each column of structured data represents a single feature. As such, they are incapable of processing unstructured text data, especially when text descriptions contain typos or inconsistent formats.
Even with structured data, one-hot encoding and manual feature crossing are needed to capture interactions, requiring enumeration of all variable combinations.
When new events are added, feature relationships must be re-maintained, creating heavy maintenance burdens as business logic evolves.
}

\rev{Counterintuitively, LLM-based knowledge reasoning is the \textit{simplest practical path} to incorporate rich domain knowledge: world knowledge comes pre-trained, the same prompt template works across countries, new promotion types require only prompt modifications, and unstructured descriptions are processed naturally without rigid parsing logic. Our contribution is not merely achieving accuracy improvements, but demonstrating a practical paradigm shift in how domain knowledge enters forecasting systems.}

\textbf{E-Commerce Forecasting with External Factors}.
To forecast dynamic e-commerce demand more reliably, recent works incorporate external business events into forecasting models. 
One direction utilizes \textit{known future information} such as planned promotions, holidays, or price changes~\cite{qi2021known,arango2025chronosx,olivares2023neural,challu2023nhits}. Qi et al. \cite{qi2021known} introduced Aliformer, integrating known upcoming promotions into Transformer self-attention. Arango et al. \cite{arango2025chronosx} propose ChronosX, which injects covariates (promotion flags, price cuts, holidays) into large pretrained time-series models. However, these techniques typically encode external inputs as structured variables, requiring complex feature engineering and failing to capture semantic richness in unstructured text. For example, two ``sales events'' may differ drastically depending on whether free shipping requires zero or 1 USD minimum spend.
Other research leverages \textit{unstructured or contextual data} about events~\cite{kalifa2022leveraging,sawhney2021fast,liu2024echo}. Kalifa et al.~\cite{kalifa2022leveraging} utilize news and Wikipedia data for anomaly periods in e-commerce, verifying the value of world event semantics for e-commerce forecasting..
Recent work has also leveraged LLMs to align time series with textual context~\cite{williams2024context,wang2024news,niu2025langtime}. 
Wang et al.~\cite{wang2024news} use LLM agents to filter and reason about relevant news articles before making time series predictions.
This underscores a key insight that beyond numeric features, textual and semantic information about interventions can greatly enhance forecast models in handling irregular scenarios. 
However, these works assume access to general-purpose texts such as news, making it hard to utilize prior business knowledge accurately.
What's more, finetuning of LLM's embeddings is usually required, incuring considerably higher computational overhead~\cite{wang2024news}.

In contrast, e-commerce demand is shaped by fast-changing business interventions whose semantic implications vary across countries and require domain-specific interpretation, stability, and high accuracy of information extraction --- characteristics existing methods do not explicitly address.

\textbf{LLMs as Time Series Forecasters}.
Recent studies explore LLMs for time series forecasting by leveraging the ability of handling sequential data and multimodal input~\cite{zhang2024large}. 
Some treat numeric sequences as token streams in zero-shot or few-shot settings~\cite{gruver2023large,hegselmann2023tabllm,tang2025time}, showing promising results on datasets with clear trends or strong seasonality. 
%
Others focus on aligning time series embeddings with language space~\cite{sun2023test,chang2023llm4ts,jin2023time}, or propose domain-specific LLMs for structured temporal tasks, such as Liu et al.~\cite{liu2023large} demonstrate LLMs are capable for health tasks with both clinical and wellness contexts, or Wang et al.~\cite{wang2024stocktime} propose StockTime for financial forecasting, where price sequences are fused with textual financial news.
%
However, empirical evidence consistently shows LLMs underperform as direct forecasters~\cite{tan2024language}. Replacing LLMs with lightweight attention layers or removing them entirely often yields comparable or better performance at a fraction of the cost, indicating LLMs are ill-suited for directly modeling temporal dynamics and dependencies~\cite{kong2025position}.

Therefore, we combine the strength of LLM-based symbolic reasoning with a lightweight attention-base forecaster. By using an LLM to convert rich unstructured event data into contextual summaries, we capture semantic nuance of intervention such as overlapping campaigns, cultural holiday significance. 

%% file: chapters/3-Methodology.tex
\section{Methodology}

\subsection{Framework Overview}

Our goal is to build a forecasting framework that is both accurate in non-stationary business settings and practical for industrial deployment. 
{\name} achieves this by decoupling semantic knowledge reasoning from numerical prediction.
The core idea is to treat the LLM as a domain-aware reasoning agent that interprets future events (e.g., campaign schedules, holidays) and converts them into interpretable textual summaries. These summaries are aligned with historical multivariate signals in a dual-tower architecture.

As shown in Figure \ref{fig:architecture}, the pipeline consists of four components:

\begin{itemize}[leftmargin=*]
    \item \textbf{Event Database and Prompting Guidance.} An operational database, maintained by business specialists, stores data about promotional campaigns, holidays, platform events, and expert knowledge. Prompting templates guide the LLM to reason about and summarize the context of a given target date.
    \item \textbf{LLM-Based KnowledgenReasoning for Future Context.} A frozen LLM processes prompts grounded in the database and outputs structured textual summaries encoding the contextual meaning of upcoming dates, which are difficult to be modeled through conventional feature engineering.
    \item \textbf{Feature Embedding and Alignment.} 
    To decouple the reasoning results from the pretrained LLM embedding space and avoid repeated heavy training, the textual summaries are tokenized into learnable embeddings, and fused with encoded historical time-series features via a fusion layer. 
    \item \textbf{Downstream Forecasting Module.} The fused representation is passed to a compact feed-forward predictor demand volumes prediction. By decoupling semantic reasoning from forecast modeling, the architecture remains scalable and efficient, while benefiting from rich future-aware signals.
\end{itemize}

This architecture offers key advantages: i) it captures rich context from unstructured business data that traditional models struggle to encode; ii) the dual-tower design ensures future event information is not diluted in the high-dimensional latent representation of historical features. This fusion also provides interpretable forecasts, allowing analysts to directly trace demand changes back to specific future events; iii) unlike approaches~\cite{jin2023time,liu2025calf,zhou2023one} that depend on LLM embeddings, our architecture uses only the textual reasoning results, making it lightweight and extensible. Stronger LLMs or agents can be seamlessly substituted without retraining. Together, these properties ensure that our method is not only accurate, but also scalable and adaptable in real-world e-commerce environments.

\subsection{Future Event Knowledge Reasoning} \label{subsec:reasoning}

Central to our framework is transforming unstructured business event data into semantically rich, human‑interpretable summaries consumable by a standard forecasting model. 

\textbf{Business Event Expert Database}. \rev{E-commerce platforms inherently require business records for campaign management, logistics planning, and seller coordination. Unlike approaches that require creating specialized knowledge bases for forecasting, we only leverage such existing operational database without additional efforts.} Specifically, Business specialists assemble a lightweight, unstructured repository of all known event interventions that materially affect demand volume, which contains: 
\begin{itemize}[leftmargin=*]
  \item \textit{Campaign Calendars:} start and end dates of platform‑wide promotions (e.g., flash sales, category discounts, free‑shipping events), along with campaign intensity levels.
  \item \textit{Holiday Schedules:} country‑specific calendars of public and religious holidays, including multi‑day festival spans.
  \item \textit{Incentive Rules:} business logic for seller‑ and product‑level incentives that vary by country or time window.
\end{itemize}

Anonymized examples from each category are provided in Appendix~\ref{app:database_example}.


\textbf{Prompt Design for Event Reasoning}. We craft a parameterized prompt template that incorporates both the Expert Database entries and the specific target prediction date. Based on operational practice, platform free shipping campaigns, holiday status and major campaign levels exert highly significant effects on order volume. By formulating well-designed questions, LLM can efficiently mine business knowledge even in one-round dialogue, which is of great importance to the stability and efficiency required by the industry. Our prompt template is extensible and designed to include a flexible set of business information, including: 
\begin{itemize}[leftmargin=*]
  \item \textit{National Context:} country identifiers and time‐zone adjustments.
  \item \textit{Temporal Properties:} day of week, month, and quarter.
  \item \textit{Calendar Events:} whether an event is occurring, the relative day within the event period, and business blackout dates.
  \item \textit{Promotional Interventions:} promotion intensity levels, campaign types, and discount methods.
  \item \textit{Logistics Incentives:} free‐shipping eligibility, minimum‐order thresholds, tiered rebates.
\end{itemize}

\rev{Beyond basic calendar queries, the LLM provides significant value in scenarios difficult to replicate with rule-based systems:
\begin{itemize}[leftmargin=*]
    \item \textit{World knowledge for cultural nuance}. The prompt includes questions. The LLM leverages pre-trained understanding that Ramadan leans more towards shopping, while Eid al-Fitr tends to focus on staying at home with family, and can determine whether the holiday is a regional-level or national-level one.
    \item \textit{Generalization to novel events}. Questions like ``How might [NewPromotionType] combined with [Holiday] affect demand?'' allow the LLM to reason about previously unseen event combinations using general knowledge.
    \item \textit{Robustness to inconsistent data}. Real campaign descriptions contain typos, abbreviations, and inconsistent formatting (e.g., ``B2G1 free!! excl. premium brands; prime only''). LLMs handle such ambiguity gracefully without brittle parsing rules.
\end{itemize}
}

This extensible design enables the system to automatically process business information for any target date without manual review or complex feature engineering, simply by adding or modifying a question in the template. An example prompt is provided in Appendix~\ref{app:prompt_reasoning_example}.
%
By instructing the LLM to reason step-by-step and emit the final structured summary within `<result>' tags, we enforce consistency and simplify parsing. We also parameterize for country-specific variations such as different holiday spans or campaign rules, allowing a single template to generalize across markets.

\textbf{Semantic Interpretation via LLM}. Based on the Expert Database and guided by the prompt, the LLM produces a structured reasoning trace and final summary.
The reasoning trace is logged for auditability but not directly used in forecasting; the summary is parsed into textual features. An example reasoning output is provided in Appendix~\ref{app:prompt_reasoning_example}.
%
To convert reasoning results to textual features, we locate the `<result>...<\/result>' segment and parse its semicolon-separated fields.
These summaries serve as \textit{textual features} encoding the business context of each forecast date. Through this design, the LLM leverages its reasoning capability, pre-trained world knowledge, and the expert database to produce high-fidelity business knowledge, far richer and more flexible than manual flags or one-hot encodings.



\begin{figure}[t!]
    \centering
    \includegraphics[width=0.79\linewidth]{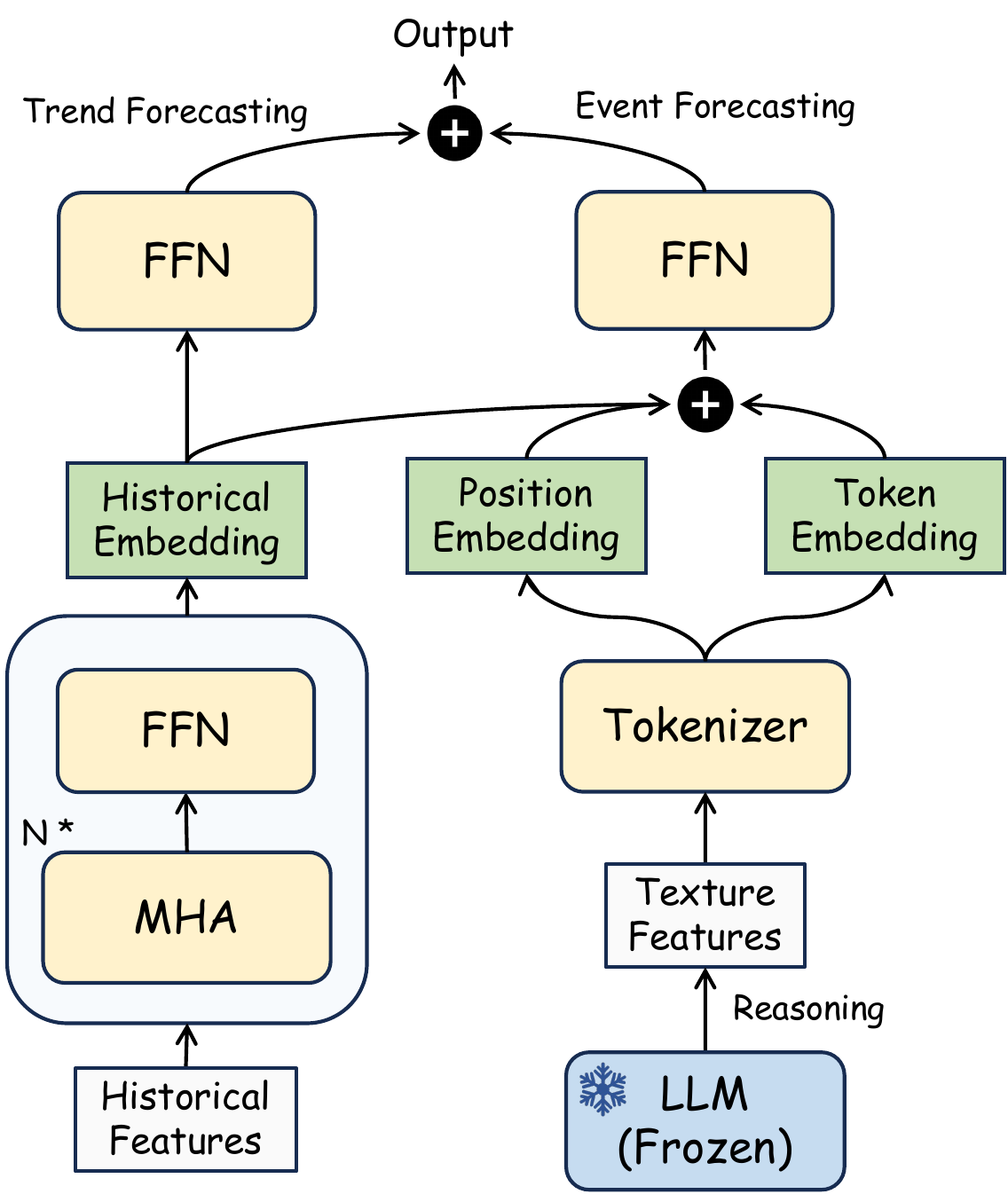}
    \vspace{-1em}
    \caption{Illustration of feature alignment and fusion.}
    \label{fig:model_overview}
\end{figure}

\subsection{Embedding and Alignment} \label{subsec:embed}

After obtaining textual reasoning outputs from the LLM, this section elaborates on embedding, aligning, and integrating reasoning features and historical data features to generate forecasts. 
Specifically, the reasoning features are tokenized and mapped into learnable token and positional embeddings, 
and aligned through a lightweight embedding alignment layer.
In parallel, historical multivariate features are encoded by a transformer encoder consisting of multi-head attention and feed-forward layers.
Then, the dual-tower architecture fuses the aligned future-event embeddings with the historical embeddings, ensuring that future event information is preserved rather than diluted in historical multivariate high-dimensional latent states.
Finally, the fused representation is passed to two forecasting heads: a trend forecasting head and an event forecasting head, whose outputs are combined to produce the final demand forecast.
Figure \ref{fig:model_overview} illustrate the structure overview.

\textbf{Historical Feature Embeddings}.
Formally, let $\mathbf{X}\in\mathbb{R}^{T\times d}$ represent the historical feature input sequence, where $T$ denotes the input length and $d$ the number of features. 
To capture cross-variate dependencies, we treat each feature’s time series as a separate token by transposing the input matrix to $\mathbf{X}_\mathrm{inv} = \mathbf{X}^\top \in \mathbb{R}^{d\times T}$, as suggested in \cite{liu2023itransformer,gao2025client}. We then feed $\mathbf{X}_\mathrm{inv}$ into a multi-head self-attention encoder to learn interactions across variables. 

Within each attention head, we first compute query $\mathbf{Q}=\mathbf{X}_\mathrm{inv} \mathbf{W}^Q$, key $\mathbf{K}=\mathbf{X}_\mathrm{inv} \mathbf{W}^K$, and value matrices $\mathbf{V}=\mathbf{X}_\mathrm{inv} \mathbf{W}^V$ via learned linear projections with $\mathbf{W}^Q, \mathbf{W}^K, \mathbf{W}^V \in \mathbb{R}^{T\times d_k}$ and $d_k$ the dimensionality of each head. Using these, we compute scaled dot-product attention across variables:
\begin{equation}
    \text{Attention}(\mathbf{Q}, \mathbf{K}, \mathbf{V}) = \text{Softmax}\left(\frac{\mathbf{Q} \mathbf{K}^\top}{\sqrt{d_k}}\right)\mathbf{V},
\end{equation}
which produces an output for each variable by weighting information from all other variables. We employ $h$ heads in parallel, the multi-head attention output is then formed by concatenating all heads and projecting back to the feature space.
This self-attention encoder \rev{yields} a final encoded \rev{vector} $\mathbf{\rev{h}} \in \mathbb{R}^{d_m \rev{\times 1}}$, where $d_m$ is the dimensionality of latent representation. \rev{Vector}
$\mathbf{\rev{h}}$ represents the target variable embedding that is learned from the multivariables.

To align the historical representation of the target variable with the dimension of the text modality, we employ a linear mapping:
\begin{equation}
    \mathbf{h}_{\text{hist}} = \mathbf{W}_{\text{proj}}^{\top}\mathbf{\rev{h}},
\end{equation}
where $\mathbf{W}_{\text{proj}} \in \mathbb{R}^{d_m \times d_{\text{align}}}$. Here, $d_{\text{align}}$ denotes the aligned embedding dimension. The resulting $\mathbf{h}_{\text{hist}}$ serves as a compact summary vector, encapsulating historical trends and cross-variate interactions in a form compatible with downstream fusion with future-event embeddings.

\textbf{Textual Feature Embedding}.
To incorporate explicit knowledge about future events, we leverage a pretrained LLM as a reasoning agent. Given a curated expert database $M$ which contains unstructured event data, we generate a dynamic textual prompt $P(M, d)$ for each forecast date $d$. 
This prompt explicitly instructs the LLM to reason about relevant contextual attributes for the given date. Formally, the LLM generates a textual reasoning response $R = \text{LLM}(P(M,d))$. Details are discussed in Section \ref{subsec:reasoning}.

We then extract a structured summary $S$ from $R$, enclosed within predefined delimiters (\texttt{<result>...</result>}), giving semantic attributes: $S = (s_1, s_2, \dots, s_k)$.
Each semantic attribute in $S$ is tokenized, and a learnable embedding is randomly initialized for each token. The learnable positional embeddings are then added to retain the sequential structure:
\begin{equation}
    \mathbf{E}_{\text{sem}}^{(i)} = \text{TokenEmbed}(s_i) + \text{PosEmbed}(i).
\end{equation}
These embeddings are aggregated into a single semantic vector: 
\begin{equation}
    \mathbf{h}_{\text{sem}} = \sum_{i=1}^{k}\mathbf{E}_{\text{sem}}^{(i)}.
\end{equation}

This embedding explicitly encodes the semantic event-context derived from LLM reasoning. 
By decoupling the reasoning results from the LLM’s embedding space and representing them through lightweight and learnable embeddings, our framework achieves both efficiency and accuracy, while preserving the ability to easily switch to stronger LLMs or agents for improved reasoning quality.

\textbf{Alignment and Forcasting}.
To combine embeddings of historical feature and reasoning feature, we first align them within a common embedding space using a linear mapping: 
$
    \mathbf{h}_{\text{aligned}} = \mathbf{h}_{\text{hist}} + \mathbf{h}_{\text{sem}}.
$
The fused embedding $\mathbf{h}_{\text{aligned}}$ is then passed into a residual-connected feed-forward network (FFN), which models nonlinear interactions while maintaining stable optimization. Each layer is defined as:
$
    \mathbf{z}^{(l+1)} = \text{LeakyReLU}\left(\text{BN}\left(\mathbf{z}^{(l)} + \text{Dropout}\left(\mathbf{W}^{(l)}\mathbf{z}^{(l)}\right)\right)\right),
$
where $\mathbf{W}^{(l)}$ represents learned parameters at layer $l$, and BN is batch normalization. 

After $L$ layers, the final hidden state $\mathbf{z}^{(L)}$ is used to produce the event-driven adjustment:
$
    \hat{y}_{\text{event}} = \mathbf{W}_{\text{event}}\mathbf{z}_{\text{event}}^{(L)} + b_{\text{event}},
$
with $\mathbf{W}_{\text{event}}, b_{\text{event}}$ as learnable projection parameters.
In parallel, the framework explicitly models the underlying demand trend from historical embeddings:
$
    \hat{y}_{\text{trend}} = \mathbf{W}_{\text{trend}}\mathbf{z}_{\text{trend}}^{(L)} + b_{\text{trend}}.
$

Finally, the total forecast is obtained by combining these two interpretable components:
\begin{equation} \label{eq:sum}
    \hat{y} = \lambda \cdot \hat{y}_{\text{trend}} + (1-\lambda) \cdot \hat{y}_{\text{event}},
\end{equation}
where $\lambda \in [0,1]$ is the weight to control the contribution of trend and event predictions, which is set to 0.4 in our model. 

One notable advantage of our framework lies in its dual-tower architecture, which decomposes the forecast into two interpretable components: the baseline trend predicted from historical data and the adjustments contributed by future business activities. This separation not only improves interpretability for practitioners but also ensures that future event information is preserved rather than diluted in the high-dimensional latent space of historical features. 

%% file: chapters/4-Experiments.tex
\begin{table*}[htbp!]
\centering
\caption{Forecasting Performance (Overall). The best performance is in \textcolor{red}{red}, and the second-best is in \textcolor{blue}{blue}.}
\vspace{-1em}
\label{tab:results_all_countries}
\resizebox{0.85\textwidth}{!}{
    \begin{tabular}{cc|cc|cc|cc|cc|cc|cc|cc}
        \toprule
            \multirow{2}{*}{Ctr.} & \multirow{2}{*}{Horiz.} & \multicolumn{2}{c|}{{\name} (Ours)} & \multicolumn{2}{c|}{Time-LLM} & \multicolumn{2}{c|}{OneFitsAll} & \multicolumn{2}{c|}{iTransformer} & \multicolumn{2}{c|}{PatchTST} & \multicolumn{2}{c|}{ChronosX} & \multicolumn{2}{c}{NeuralProphet}\\
            \cmidrule(lr){3-4} \cmidrule(lr){5-6} \cmidrule(lr){7-8} \cmidrule(lr){9-10} \cmidrule(lr){11-12} \cmidrule(lr){13-14} \cmidrule(lr){15-16}
            & & MAE & MSE & MAE & MSE & MAE & MSE & MAE & MSE & MAE & MSE & MAE & MSE & MAE & MSE\\
        \midrule
            \multirow{4}{*}{01}
            & 1 & \textcolor{red}{\textbf{0.270}} & \textcolor{red}{\textbf{0.123}} & \textcolor{blue}{0.346} & \textcolor{blue}{0.208} & 0.561 & 0.728 & 0.469 & 0.505 & 0.849 & 1.662 & 0.416 & 0.253 & 0.641 & 0.594 \\
            & 2 & \textcolor{red}{\textbf{0.288}} & \textcolor{red}{\textbf{0.144}} & 0.516 & 0.470 & 0.644 & 0.997 & 0.688 & 1.123 & 0.888 & 1.865 & \textcolor{blue}{0.496} & \textcolor{blue}{0.369} & 0.653 & 0.616 \\
            & 3 & \textcolor{red}{\textbf{0.312}} & \textcolor{red}{\textbf{0.177}} & 0.611 & 0.779 & 0.762 & 1.271 & 0.779 & 1.389 & 0.919 & 2.037 & \textcolor{blue}{0.556} & \textcolor{blue}{0.480} & 0.665 & 0.735 \\
            & 4 & \textcolor{red}{\textbf{0.337}} & \textcolor{red}{\textbf{0.205}} & 0.721 & 1.155 & 0.940 & 1.768 & 0.883 & 1.537 & 0.945 & 2.184 & \textcolor{blue}{0.597} & \textcolor{blue}{0.572} & 0.669 & 0.672 \\
        \midrule
            \multirow{4}{*}{02}
            & 1 & \textcolor{blue}{0.762} & \textcolor{blue}{0.899} & 0.968 & 1.924 & 1.015 & 1.783 & 1.471 & 9.475 & 2.755 & 14.385 & \textcolor{red}{\textbf{0.691}} & \textcolor{red}{\textbf{0.679}} & 1.623 & 4.894 \\
            & 2 & \textcolor{red}{\textbf{0.786}} & \textcolor{red}{\textbf{0.944}} & 1.222 & 3.512 & 1.274 & 2.672 & 1.522 & 3.884 & 2.964 & 16.288 & \textcolor{blue}{0.840} & \textcolor{blue}{1.065} & 1.845 & 14.198 \\
            & 3 & \textcolor{red}{\textbf{0.801}} & \textcolor{red}{\textbf{0.948}} & 1.653 & 5.856 & 1.462 & 3.536 & 1.941 & 6.121 & 3.162 & 18.104 & \textcolor{blue}{0.970} & \textcolor{blue}{1.505} & 1.707 & 8.385 \\
            & 4 & \textcolor{red}{\textbf{0.831}} & \textcolor{red}{\textbf{0.983}} & 1.913 & 8.132 & 1.803 & 5.579 & 1.930 & 6.396 & 3.312 & 19.678 & \textcolor{blue}{1.084} & \textcolor{blue}{2.065} & 1.909 & 16.530 \\
        \midrule
            \multirow{4}{*}{03}
            & 1 & \textcolor{blue}{1.616} & 4.571 & 1.940 & 6.591 & 1.672 & \textcolor{blue}{4.422} & 2.622 & 11.584 & 3.400 & 28.432 & \textcolor{red}{\textbf{1.446}} & \textcolor{red}{\textbf{2.817}} & 1.973 & 6.507 \\
            & 2 & \textcolor{red}{\textbf{1.652}} & \textcolor{red}{\textbf{4.346}} & 2.363 & 10.259 & 2.115 & 7.521 & 3.398 & 28.006 & 3.619 & 31.970 & \textcolor{blue}{1.742} & \textcolor{blue}{4.839} & 2.151 & 14.864 \\
            & 3 & \textcolor{red}{\textbf{1.647}} & \textcolor{red}{\textbf{4.780}} & 2.584 & 12.510 & 2.622 & 11.464 & 2.996 & 25.723 & 3.890 & 35.793 & \textcolor{blue}{1.815} & \textcolor{blue}{5.051} & 2.150 & 10.858 \\
            & 4 & \textcolor{red}{\textbf{1.694}} & \textcolor{red}{\textbf{4.826}} & 3.012 & 17.172 & 3.369 & 18.547 & 3.274 & 21.452 & 4.040 & 37.536 & \textcolor{blue}{1.734} & \textcolor{blue}{4.987} & 2.155 & 9.470 \\
        \midrule
            \multirow{4}{*}{04}
            & 1 & \textcolor{red}{\textbf{0.718}} & \textcolor{red}{\textbf{0.753}} & 0.792 & 1.096 & 1.017 & 1.607 & \textcolor{blue}{0.733} & \textcolor{blue}{0.841} & 1.229 & 2.594 & 0.802 & 1.033 & 0.936 & 1.401 \\
            & 2 & \textcolor{red}{\textbf{0.675}} & \textcolor{red}{\textbf{0.698}} & 0.982 & 1.659 & 1.196 & 2.653 & 0.950 & 1.453 & 1.307 & 2.842 & \textcolor{blue}{0.786} & \textcolor{blue}{1.006} & 0.933 & 1.403 \\
            & 3 & \textcolor{red}{\textbf{0.702}} & \textcolor{red}{\textbf{0.732}} & 1.090 & 2.042 & 1.453 & 3.642 & 1.198 & 2.397 & 1.374 & 3.063 & \textcolor{blue}{0.795} & \textcolor{blue}{0.980} & 0.941 & 1.417 \\
            & 4 & \textcolor{red}{\textbf{0.680}} & \textcolor{red}{\textbf{0.709}} & 1.196 & 2.523 & 1.513 & 3.668 & 1.669 & 4.195 & 1.432 & 3.255 & \textcolor{blue}{{0.768}} & \textcolor{blue}{0.944} & 0.952 & 1.450 \\
        \midrule
            \multicolumn{2}{c|}{Average} & \textcolor{red}{\textbf{0.885}} & \textcolor{red}{\textbf{1.682}} & 1.370 & 4.743 & 1.464 & 4.491 & 1.658 & 7.880 & 2.255 & 14.105 & \textcolor{blue}{0.948} & \textcolor{blue}{1.723} & 1.369 & 5.874 \\
        \bottomrule
    \end{tabular}
}
\end{table*}

\begin{table*}[htbp!]
\centering
\caption{Forecasting Performance (Event-Driven Periods). The best performance is in \textcolor{red}{red}, and the second-best is in \textcolor{blue}{blue}.}
\vspace{-1em}
\label{tab:results_event_driven}
\resizebox{0.85\textwidth}{!}{
    \begin{tabular}{cc|cc|cc|cc|cc|cc|cc|cc}
        \toprule
            \multirow{2}{*}{Ctr.} & \multirow{2}{*}{Horiz.} & \multicolumn{2}{c|}{{\name} (Ours)} & \multicolumn{2}{c|}{Time-LLM} & \multicolumn{2}{c|}{OneFitsAll} & \multicolumn{2}{c|}{iTransformer} & \multicolumn{2}{c|}{PatchTST} & \multicolumn{2}{c|}{ChronosX} & \multicolumn{2}{c}{NeuralProphet}\\
            \cmidrule(lr){3-4} \cmidrule(lr){5-6} \cmidrule(lr){7-8} \cmidrule(lr){9-10} \cmidrule(lr){11-12} \cmidrule(lr){13-14} \cmidrule(lr){15-16}
            & & MAE & MSE & MAE & MSE & MAE & MSE & MAE & MSE & MAE & MSE & MAE & MSE & MAE & MSE\\
        \midrule
            \multirow{4}{*}{01}
            & 1 & \textcolor{red}{\textbf{0.368}} & \textcolor{red}{\textbf{0.205}} & 0.489 & 0.343 & 0.476 & 0.324 & 0.424 & 0.276 & 2.722 & 7.849 & \textcolor{blue}{0.375} & \textcolor{blue}{0.221} & 0.565 & 0.374 \\
            & 2 & \textcolor{red}{\textbf{0.334}} & \textcolor{red}{\textbf{0.181}} & 0.898 & 1.023 & 0.535 & 0.477 & 0.613 & 0.669 & 2.900 & 8.895 & \textcolor{blue}{0.396} & \textcolor{blue}{0.255} & 0.559 & 0.369 \\
            & 3 & \textcolor{red}{\textbf{0.282}} & \textcolor{red}{\textbf{0.132}} & 1.428 & 2.578 & 1.017 & 1.498 & 0.739 & 0.823 & 3.037 & 9.742 & \textcolor{blue}{0.360} & \textcolor{blue}{0.215} & 0.630 & 1.107 \\
            & 4 & \textcolor{red}{\textbf{0.293}} & \textcolor{red}{\textbf{0.137}} & 2.026 & 4.685 & 1.645 & 3.584 & 1.040 & 1.485 & 3.144 & 10.433 & \textcolor{blue}{0.399} & \textcolor{blue}{0.239} & 0.580 & 0.563 \\
        \midrule
            \multirow{4}{*}{02}
            & 1 & \textcolor{red}{\textbf{0.358}} & \textcolor{red}{\textbf{0.194}} & 1.797 & 4.753 & 0.989 & 1.387 & 1.112 & 1.731 & 6.806 & 47.951 & \textcolor{blue}{0.638} & \textcolor{blue}{0.604} & 0.849 & 6.729 \\
            & 2 & \textcolor{red}{\textbf{0.483}} & \textcolor{red}{\textbf{0.337}} & 2.551 & 10.281 & 1.464 & 2.789 & 1.564 & 3.065 & 7.253 & 54.090 & \textcolor{blue}{1.028} & \textcolor{blue}{1.589} & 1.589 & 44.220 \\
            & 3 & \textcolor{red}{\textbf{0.631}} & \textcolor{red}{\textbf{0.533}} & 3.472 & 18.073 & 1.754 & 4.701 & 2.548 & 9.741 & 7.609 & 59.411 & \textcolor{blue}{1.519} & \textcolor{blue}{3.256} & 0.923 & 19.300 \\
            & 4 & \textcolor{red}{\textbf{0.782}} & \textcolor{red}{\textbf{0.759}} & 4.378 & 26.717 & 2.442 & 7.693 & 3.180 & 15.212 & 7.898 & 64.012 & \textcolor{blue}{2.062} & \textcolor{blue}{5.471} & 1.530 & 56.255 \\
        \midrule
            \multirow{4}{*}{03}
            & 1 & \textcolor{red}{\textbf{1.409}} & \textcolor{red}{\textbf{3.164}} & 5.244 & 29.015 & 2.527 & 9.595 & 3.853 & 20.542 & 12.876 & 168.550 & \textcolor{blue}{1.755} & \textcolor{blue}{4.174} & 3.550 & 17.610 \\
            & 2 & \textcolor{red}{\textbf{1.128}} & \textcolor{red}{\textbf{2.228}} & 7.013 & 51.710 & 1.747 & 4.355 & 2.850 & 9.176 & 13.573 & 187.378 & \textcolor{blue}{2.094} & \textcolor{blue}{6.037} & 4.650 & 83.458 \\
            & 3 & \textcolor{red}{\textbf{0.978}} & \textcolor{red}{\textbf{1.732}} & 7.393 & 58.916 & 3.242 & 17.229 & \textcolor{blue}{2.152} & \textcolor{blue}{5.735} & 14.132 & 203.262 & 2.392 & 7.879 & 4.666 & 51.045 \\
            & 4 & \textcolor{red}{\textbf{1.074}} & \textcolor{red}{\textbf{1.940}} & 8.889 & 82.473 & 5.762 & 42.985 & 2.990 & 10.905 & 14.335 & 209.261 & \textcolor{blue}{2.102} & \textcolor{blue}{6.731} & 4.275 & 37.635 \\
        \midrule
            \multirow{4}{*}{04}
            & 1 & \textcolor{red}{\textbf{0.555}} & \textcolor{red}{\textbf{0.472}} & 1.906 & 4.054 & 0.798 & 0.801 & \textcolor{blue}{0.587} & \textcolor{blue}{0.486} & 3.384 & 11.944 & 0.651 & 0.685 & 2.082 & 4.679 \\
            & 2 & \textcolor{red}{\textbf{0.447}} & \textcolor{red}{\textbf{0.321}} & 2.554 & 7.070 & \textcolor{blue}{0.449} & \textcolor{blue}{{0.323}} & 0.469 & 0.391 & 3.435 & 12.324 & 0.716 & 0.903 & 2.069 & 4.637 \\
            & 3 & \textcolor{red}{\textbf{0.436}} & \textcolor{red}{\textbf{0.293}} & 2.945 & 9.103 & 0.583 & \textcolor{blue}{0.526} & \textcolor{blue}{0.568} & 0.595 & 3.450 & 12.408 & 0.740 & 0.838 & 2.067 & 4.635 \\
            & 4 & \textcolor{red}{\textbf{0.347}} & \textcolor{red}{\textbf{0.191}} & 3.285 & 11.264 & 1.467 & 2.969 & \textcolor{blue}{0.821} & \textcolor{blue}{1.203} & 3.440 & 12.339 & 0.731 & 0.839 & 2.080 & 4.682 \\
        \midrule
            \multicolumn{2}{c|}{Average} & \textcolor{red}{\textbf{0.619}} & \textcolor{red}{\textbf{0.802}} & 3.517 & 20.129 & 1.681 & 6.328 & 1.595 & 5.066 & 6.875 & 67.491 & \textcolor{blue}{1.123} & \textcolor{blue}{2.496} & 2.041 & 21.012 \\
        \bottomrule
    \end{tabular}
}
\end{table*}

\begin{figure*}[t]%
    \centering
    \begin{subfigure}{0.245\linewidth}
        \includegraphics[width=\linewidth]{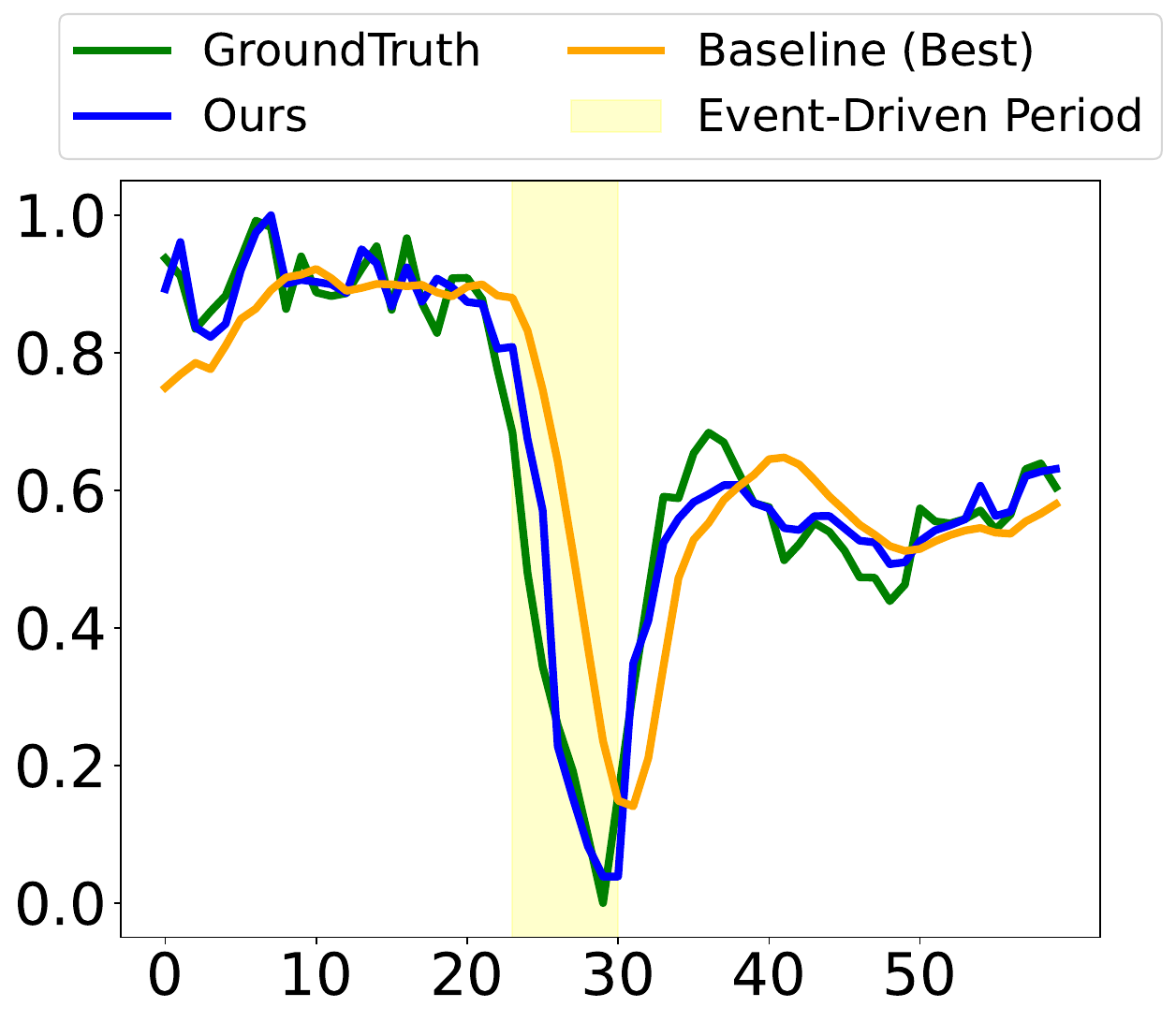}
        \caption{Country 01.}
    \end{subfigure}
    \begin{subfigure}{0.245\linewidth}
        \includegraphics[width=\linewidth]{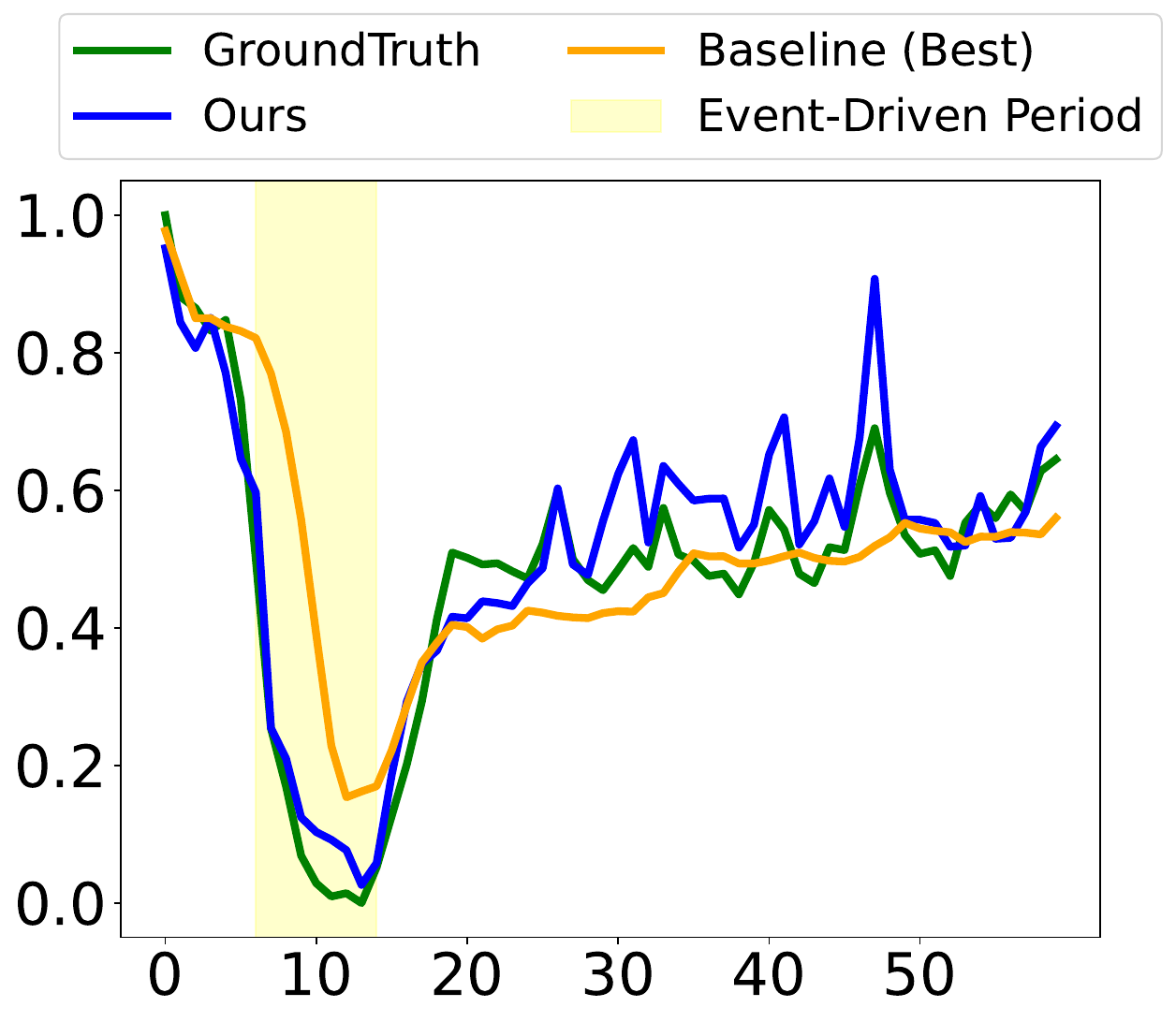}
        \caption{Country 02.}
    \end{subfigure}
    \begin{subfigure}{0.245\linewidth}
        \includegraphics[width=\linewidth]{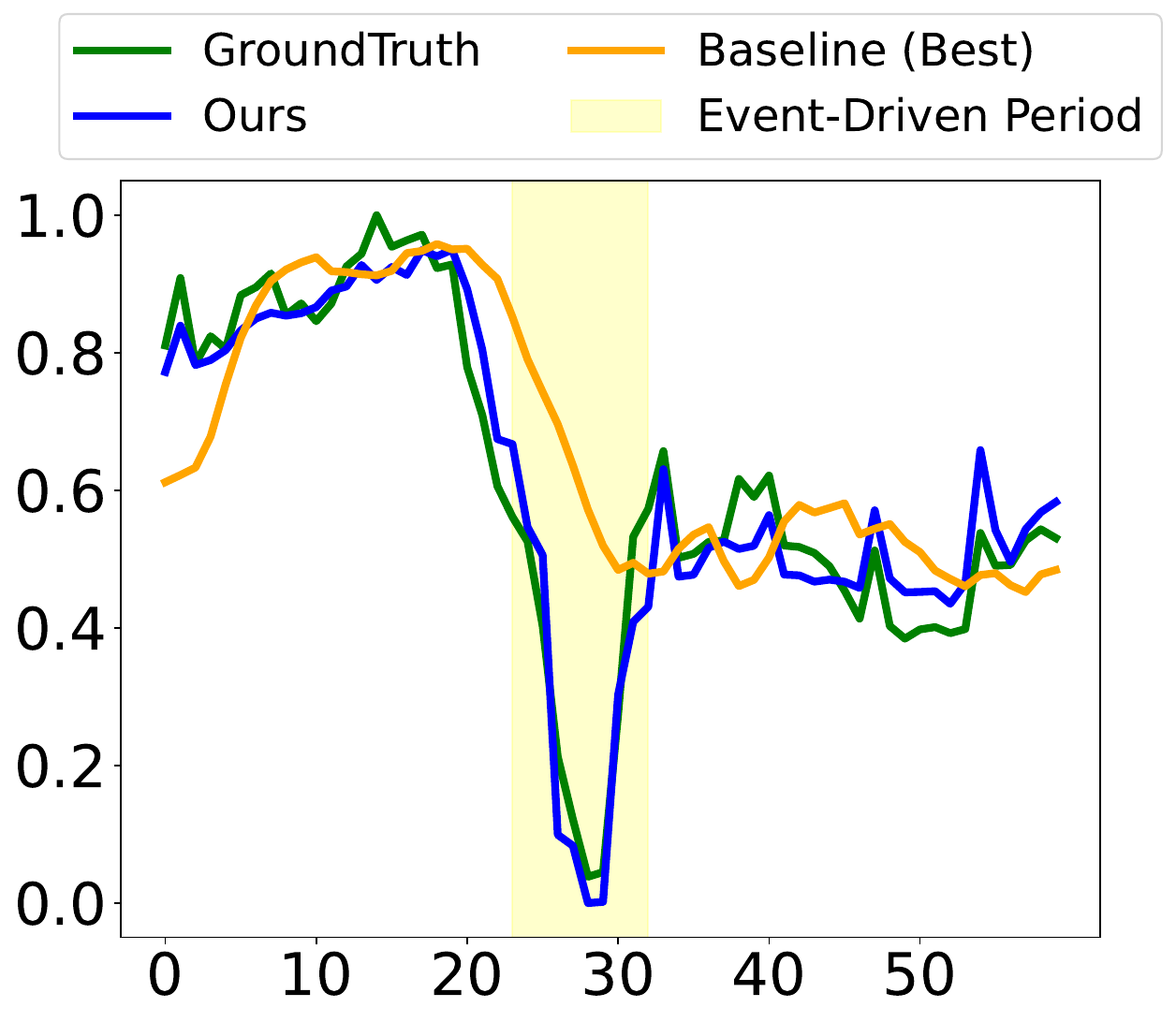}
        \caption{Country 03.}
    \end{subfigure}
    \begin{subfigure}{0.245\linewidth}
        \includegraphics[width=\linewidth]{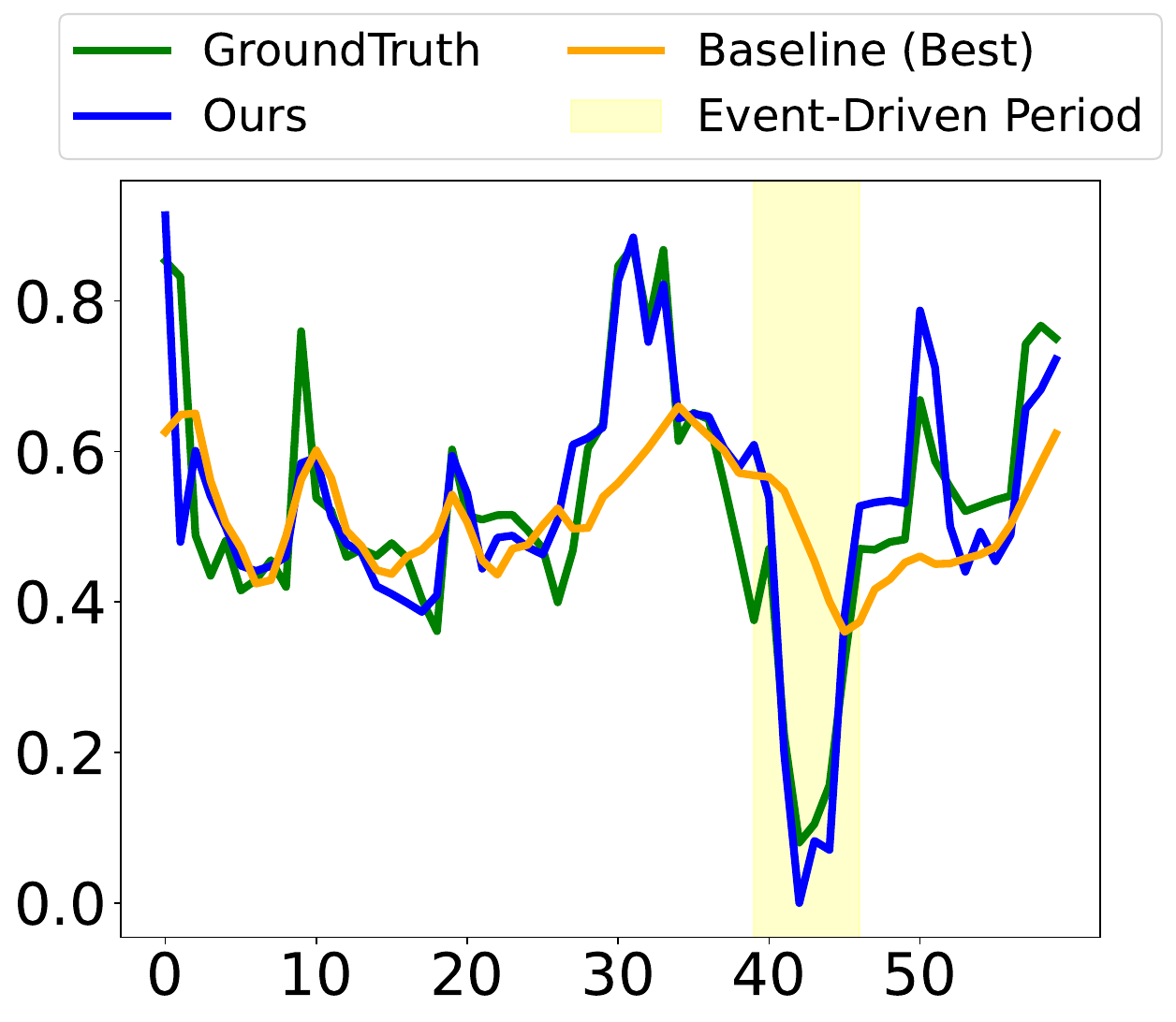}
        \caption{Country 04.}
    \end{subfigure}
    \vspace{-0.9em}
    \caption{The visualization of normalized prediction results. The \hl{highlighted} area corresponds to a major holiday sales event. We present predictions from our method (in \textcolor{blue}{blue}), the best baseline (in \textcolor{orange}{orange}), and the ground truth (in \textcolor{green}{green}). 
    }%
    \label{fig:holiday_comparison}
\end{figure*}

\section{Experiments}

\subsection{Experimental Settings}

\textbf{Datasets}. 
We deploy on a major e-commerce platform spanning January--August 2025 across 4 countries and 160 regions, comprising over 90,000 \rev{daily-granularity} records. All volumes are normalized and dates/regions are anonymized. 
%

\rev{
\textbf{Event Statistics and Impact}. 
We classified major events into: i) promotional campaigns (category discounts, seller incentives, free shipping), accounting for 30\% of the total; ii) religious/cultural/ public holidays, accounting for 15\% of the total. About 38\% of event-driven days involve two or more simultaneous events, with the most common overlaps being free shipping campaigns and seller incentives. 
Promotional campaigns show average demand increases of 85--120\% during peak periods; religious/cultural holidays exhibit biphasic patterns (surge before, drop during); public holidays show 30--50\% reductions due to travel and leisure activities
}

\textbf{Metrics}. 
We adopt MSE and MAE, evaluated with a 14-day look-back window to forecast the next 1--4 days (T+1 to T+4). This reflects common e-commerce practice where detailed promotional plans are finalized shortly before execution, making 1--4 day horizons most actionable
Longer horizons carry greater strategic value for operational management, and the framework can extend to longer horizons when richer future information is available.

\textbf{Comparison Methods}. 
\rev{We compare against a diverse set of state-of-the-art models: i) Attention-based methods: 
PatchTST \cite{nie2022time}, iTransformer \cite{liu2023itransformer}, 
; 
ii) Large-Model-based methods: Time-LLM~\cite{jin2023time}, OneFitsAll~\cite{zhou2023one}, and ChronosX~\cite{ArangoMKASSSTSMBWR25}; and iii) Statistical/Hybrid method: NeuralProphet~\cite{triebe2021neuralprophet}.}
These baselines span a range of architectural paradigms, ensuring a comprehensive evaluation.

\rev{All baseline methods receive the same LLM-reasoned event information. For attention-based (PatchTST, iTransformer) and LLM-based models (Time-LLM, OneFitsAll), we encode the LLM-generated event summaries just as {\name} does, and concatenate the resulting embeddings with the historical time series features as additional input channels. 
For ChronosX, which supports string covariate inputs, we manually parse multiple events from the summary text, with each event treated as a separate feature column.
For NeuralProphet, we encode text as one-hot categorical features.
}

\textbf{Implementation.}
Baselines (except LLM-based methods) are implemented with TSLib~\cite{wang2024tssurvey}
using the optimized hyperparameters on NVIDIA H100 GPUs with L2 loss.
Our model uses PyTorch. Reasoning features are obtained by querying a frozen LLM, tokenized using GLM's tokenizer~\cite{glm2024chatglm}. 
Token and positional embeddings are randomly initialized, summed, and aggregated. Both historical and reasoning embeddings are projected into a 1024-dimensional shared latent space. 
Only the encoder and forecasting layers are trainable.

\textbf{Deployment.}
{\name} has been deployed in production on a major e-commerce platform since March 2025, retrained weekly on the latest 13 months of data. 
Daily forecasts (T+1 to T+4) support inventory allocation, warehouse staffing, and campaign preparation. For instance, warehouse managers receive 4-day forecasts every morning to adjust manpower planning, while category managers use the event-sensitive forecasts to anticipate demand surges during campaign days.
With approximately 150M parameters, training completes within 4 minutes/epoch and inference under 20 seconds.
%


\subsection{Results}

\textbf{Overall Performance.}
Table~\ref{tab:results_all_countries} presents the overall results. 
{\name} achieves the best average performance with MAE of 0.885 and MSE of 1.682, outperforming the strongest baseline ChronosX (MAE=0.948, MSE=1.723) by {6.65\%} in MAE and {2.38\%} in MSE. Compared to other baselines, the improvements are more substantial: \textbf{35.4\%} MAE and \textbf{64.5\%} MSE reduction versus Time-LLM, \textbf{39.5\%} MAE and \textbf{62.5\%} MSE reduction versus OneFitsAll, and \textbf{46.6\%} MAE and \textbf{78.7\%} MSE reduction versus iTransformer.

For Country 01, {\name} dominates across all horizons, achieving an average improvement of {41.5\%} in MAE and {61.3\%} in MSE compared to ChronosX. In Country 02, while ChronosX achieves the best performance at T+1, {\name} consistently outperforms all baselines at longer horizons (T+2 to T+4), with overall improvements of {11.3\%} MAE and {29.0\%} MSE. Country 03 presents the most competitive scenario, where ChronosX leads at T+1 and achieves superior MSE at several horizons; however, {\name} still achieves the best MAE at T+2, T+3, and T+4, demonstrating robustness at longer prediction horizons where uncertainty increases. For Country 04, {\name} achieves consistent superiority across all horizons, reducing MAE by {11.9\%} and MSE by {27.0\%} compared to ChronosX.


\textbf{Event-Driven Periods.}
The results presented in Table~\ref{tab:results_event_driven} specifically illustrate model performance during critical event-driven periods.
During these intervals, demand exhibits heightened volatility and deviation from historical patterns, challenging forecasting models considerably. 
{\name} significantly outperforms all baselines, achieving an average MAE of 0.619 and MSE of 0.802, compared to the strongest baseline ChronosX (MAE=1.123, MSE=2.496), representing improvements of \textbf{44.9\%} in MAE and \textbf{67.9\%} in MSE. 

The performance gap widens substantially compared to overall performance, highlighting {\name}'s strength in capturing event-driven demand dynamics.
For Country 01, {\name} achieves consistent dominance across all horizons, with average improvements of \textbf{16.7\%} in MAE and \textbf{29.6\%} in MSE over ChronosX. Country 02 demonstrates the most pronounced improvements, where {\name} reduces MAE by \textbf{57.0\%} and MSE by \textbf{83.3\%} compared to ChronosX, reflecting the model's ability to capture volatile promotional effects.
Country 03, which exhibits the highest absolute demand volatility, still shows substantial improvements of \textbf{45.0\%} in MAE and \textbf{63.5\%} in MSE.
In Country 04, {\name} achieves improvements of \textbf{37.2\%} in MAE and \textbf{60.9\%} in MSE, maintaining consistent superiority across all horizons.

To better illustrate the performance of our method, we show the visualization of the predictions over a 60-day timeframe at T+4 of our method and the best baseline model in Figure \ref{fig:holiday_comparison}
The highlighted event-driven period, encompassing holidays and promotional campaigns, exhibits pronounced fluctuations in demand volume. Our model (blue line) closely tracks the ground truth, capturing the sharp declines and subsequent recoveries in demand volumes more accurately than the best baseline method. 
These observations highlight the importance and practical benefit of incorporating event reasoning provided by LLMs, resulting in superior forecasting performance during critical business periods.


\textbf{LLM for knowledge reasoning vs. LLM for forecasting}.
A critical design question in LLM-augmented forecasting is \textit{how} to leverage LLM capabilities: should LLMs directly participate in numerical prediction, or should they serve as reasoning engines that extract semantic knowledge for downstream forecasters? Our experiments provide strong evidence for the latter.

We compare {\name} against two representative LLM-based forecasting methods: Time-LLM, which reprograms time series as text prompts for LLM processing, and OneFitsAll, which uses frozen LLM embeddings as feature extractors. As shown in Table~\ref{tab:results_all_countries}, {\name} achieves an average MAE of 0.885 and MSE of 1.682, compared to Time-LLM (MAE=1.370, MSE=4.743) and OneFitsAll (MAE=1.464, MSE=4.491). This represents improvements of \textbf{35.4\%} MAE and \textbf{64.5\%} MSE over Time-LLM, and \textbf{39.5\%} MAE and \textbf{62.5\%} MSE over OneFitsAll.
The performance gap becomes even more pronounced during event-driven periods (Table~\ref{tab:results_event_driven}), where {\name} achieves \textbf{82.4\%} lower MAE and \textbf{96.0\%} lower MSE than Time-LLM, and \textbf{63.2\%} lower MAE and \textbf{87.3\%} lower MSE than OneFitsAll. 

We attribute this to a fundamental mismatch between LLM architecture and numerical forecasting requirements. LLMs are trained via next-token prediction on discrete vocabularies, which introduces several limitations for time series: i) numerical values must be tokenized, losing precision and introducing quantization artifacts; ii) the autoregressive generation process accumulates errors across prediction horizons. 
In contrast, {\name} employs a separation of concerns principle: LLMs handle what they excel at, i.e., interpreting complex event descriptions, reasoning about business context, and generating structured semantic summaries. And a specialized forecasting module handles numerical prediction. This division allows each component to operate in its native domain, avoiding the impedance mismatch that plagues direct LLM forecasting approaches. The LLM's output serves as a semantic prior that guides the forecaster, rather than attempting to replace it.
 


\begin{table}[t]
\centering
\caption{\rev{Performance comparison w/ and w/o future event features during event-driven periods (averaged across all countries). The best three baselines are compared.}} \label{tab:ablation_reasoning}
\vspace{-1em}
\label{tab:ablation_event_features}
\resizebox{\linewidth}{!}{
\begin{tabular}{c|c|cc|cc|cc}
    \toprule
    & \multirow{2}{*}{Horiz.} & \multicolumn{2}{c|}{w/ Future Events} & \multicolumn{2}{c|}{w/o Future Events} & \multicolumn{2}{c}{Improv. (\%)} \\
    \cmidrule(lr){3-4} \cmidrule(lr){5-6} \cmidrule(lr){7-8}
    & & MAE & MSE & MAE & MSE & MAE & MSE \\
    \midrule
    \multirow{4}{*}{\rotatebox{90}{\shortstack{{\name}}}}
        & 1 & \textbf{0.673} & \textbf{1.009} & 1.551 & 4.328 & \textcolor{red}{56.60\% $\uparrow$} & \textcolor{red}{76.68\% $\uparrow$} \\
        & 2 & \textbf{0.598} & \textbf{0.767} & 1.965 & 6.538 & \textcolor{red}{69.57\% $\uparrow$} & \textcolor{red}{88.27\% $\uparrow$} \\
        & 3 & \textbf{0.582} & \textbf{0.673} & 2.779 & 12.188 & \textcolor{red}{79.05\% $\uparrow$} & \textcolor{red}{94.48\% $\uparrow$} \\
        & 4 & \textbf{0.624} & \textbf{0.757} & 3.625 & 18.832 & \textcolor{red}{82.79\% $\uparrow$} & \textcolor{red}{95.98\% $\uparrow$} \\
    \midrule
    \multirow{4}{*}{\rotatebox{90}{\shortstack{ChronosX}}}
        & 1 & \textbf{0.855} & \textbf{1.421} & 1.122 & 2.636 & \textcolor{red}{23.79\% $\uparrow$} & \textcolor{red}{46.09\% $\uparrow$} \\
        & 2 & \textbf{1.059} & \textbf{2.196} & 2.055 & 7.880 & \textcolor{red}{48.48\% $\uparrow$} & \textcolor{red}{72.13\% $\uparrow$} \\
        & 3 & \textbf{1.253} & \textbf{3.047} & 3.009 & 14.373 & \textcolor{red}{58.37\% $\uparrow$} & \textcolor{red}{78.80\% $\uparrow$} \\
        & 4 & \textbf{1.324} & \textbf{3.320} & 3.854 & 21.266 & \textcolor{red}{65.65\% $\uparrow$} & \textcolor{red}{84.39\% $\uparrow$} \\
    \midrule
    \multirow{4}{*}{\rotatebox{90}{\shortstack{OneFitsAll}}}
        & 1 & 1.198 & 3.027 & \textbf{1.100} & \textbf{2.191} & -8.18\% & -27.62\% \\
        & 2 & \textbf{1.049} & \textbf{1.986} & 1.640 & 4.567 & \textcolor{red}{36.04\% $\uparrow$} & \textcolor{red}{56.52\% $\uparrow$} \\
        & 3 & \textbf{1.649} & \textbf{5.989} & 2.435 & 9.583 & \textcolor{red}{32.28\% $\uparrow$} & \textcolor{red}{37.51\% $\uparrow$} \\
        & 4 & \textbf{2.829} & \textbf{14.308} & 3.161 & 15.107 & \textcolor{red}{10.51\% $\uparrow$} & \textcolor{red}{5.29\% $\uparrow$} \\
    \midrule
    \multirow{4}{*}{\rotatebox{90}{\shortstack{{iTransform.}}}}
        & 1 & 1.494 & 5.759 & \textbf{1.162} & \textbf{2.051} & -22.22\% & -64.39\% \\
        & 2 & \textbf{1.374} & \textbf{3.325} & 1.847 & 5.248 & \textcolor{red}{25.61\% $\uparrow$} & \textcolor{red}{36.64\% $\uparrow$} \\
        & 3 & \textbf{1.502} & \textbf{4.224} & 2.608 & 10.205 & \textcolor{red}{42.41\% $\uparrow$} & \textcolor{red}{58.60\% $\uparrow$} \\
        & 4 & \textbf{2.008} & \textbf{7.201} & 3.448 & 16.943 & \textcolor{red}{41.76\% $\uparrow$} & \textcolor{red}{57.50\% $\uparrow$} \\
    \bottomrule
\end{tabular}
}
\end{table}

\textbf{Effectiveness of Future Event Knowledge Reasoning}.
To isolate the contribution of future event reasoning, we conduct an ablation study comparing model performance with and without event features during event-driven periods. The best three baselines accorinf to Table~\ref{tab:results_event_driven} are chosen for comparison. As shown in Table~\ref{tab:ablation_reasoning}, {\name} exhibits substantial and consistent improvements when equipped with future event reasoning: \textbf{67.2\%} MAE and \textbf{84.3\%} MSE reduction at T+1, increasing to \textbf{86.9\%} MAE and \textbf{97.7\%} MSE reduction at T+4. This monotonic increase in improvement with horizon length is intuitive, that longer forecasting horizons involve greater uncertainty, and future event knowledge becomes increasingly valuable for resolving this uncertainty.

Notably, the ablation reveals divergent patterns across methods. While {\name} benefits consistently from event features at all horizons, baseline methods show mixed results. OneFitsAll and iTransformer degrade at T+1 with event features ($-8.2\%$ and $-22.2\%$ MAE respectively) but improve at longer horizons (T+2 to T+4). This suggests a fundamental limitation in how these architectures integrate heterogeneous information sources. At short horizons where historical patterns remain strongly predictive, naively concatenating event embeddings may introduce noise that overwhelms the useful historical signal. Only at longer horizons, where historical extrapolation becomes unreliable, do the event features provide net benefit. 
{\name}'s dual-encoder architecture explicitly addresses this challenge through separate processing pathways for historical patterns and event reasoning, followed by learned fusion. 
The architecture learns when to rely on historical trends versus when to prioritize event signals.

Besides, since most widely used time series models do not support the incorporation of future event information, we also evaluate the performance of these native models that rely solely on historical observations, to emphasize the importance of event knowledge for e-commerce demand forecasting. In short, {\name} outperforms all native baselines by a significant margin, with improvements of \textbf{66.7\%}/\textbf{88.2\%} during event-driven periods over the best native baseline. The detailed results and analysis can be found in Appendix~\ref{app:native_models}.



\begin{figure}[t]%
    \centering
    \begin{subfigure}{0.48\linewidth}
        \includegraphics[width=\linewidth]{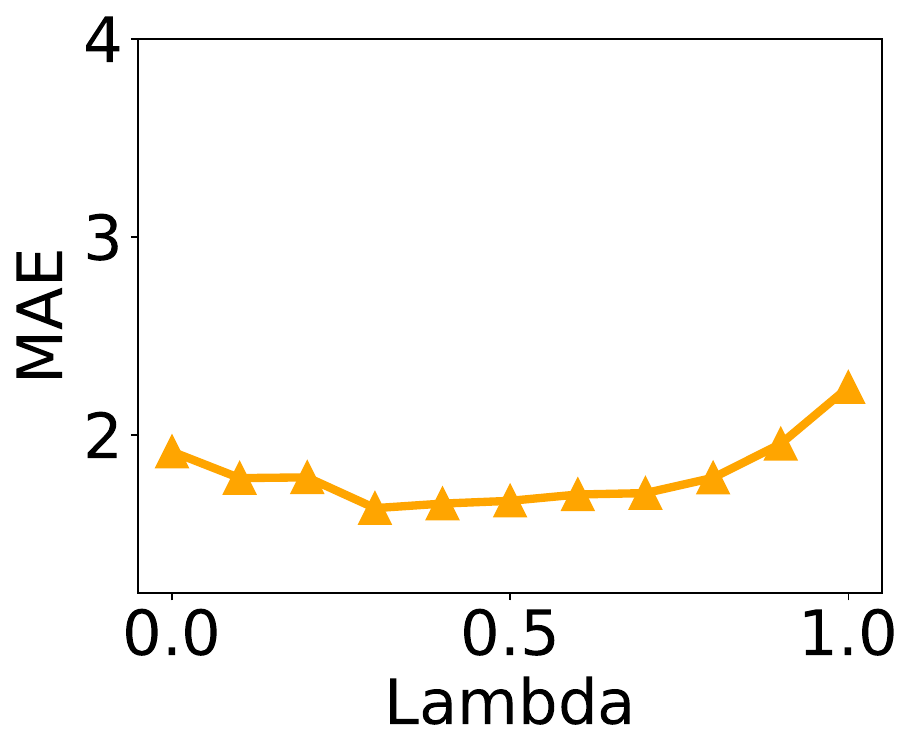}
        \caption{Overall Period.}
    \end{subfigure}
    \begin{subfigure}{0.48\linewidth}
        \includegraphics[width=\linewidth]{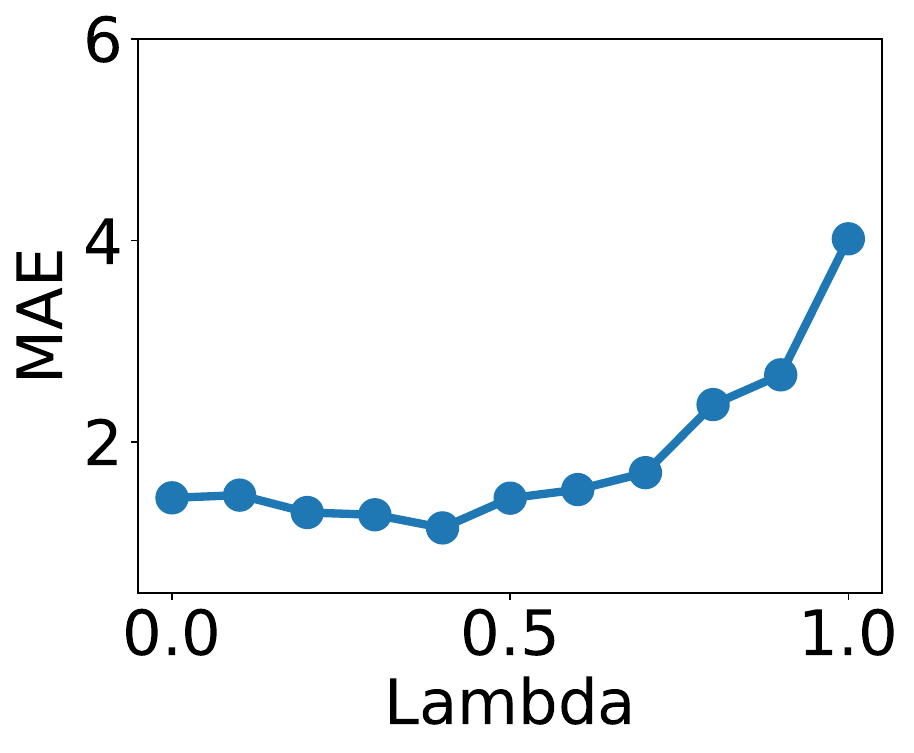}
        \caption{Event-Driven Period.}
    \end{subfigure}
    \vspace{-1.2em}
    \caption{\rev{The effect of $\lambda$ of Eq.~\ref{eq:sum} on demand forecasting.}}%
    \label{fig:lambda_analysis}
\end{figure}

\textbf{Impact of $\lambda$.}
Figure~\ref{fig:lambda_analysis} shows the effect of $\lambda$ (Eq.~\ref{eq:sum}), which balances historical and event features.
We investigate its impact by varying $\lambda$ from 0 (historical features only) to 1 (event features only) across all forecasting horizons for Country 03, as shown in Figure~\ref{fig:lambda_analysis}.
For overall performance, the optimal $\lambda$ lies around 0.3, achieving MAE of 1.629 compared to 1.915 at $\lambda=0$ ({14.9\%} improvement). For event-driven periods, the optimal shifts slightly higher to $\lambda \approx 0.4$, yielding MAE of 1.147 versus 1.446 at $\lambda=0$ ({20.6\%} improvement). This difference reflects the intuition that the contribution of trend and event predictions can be controlled by setting an appropriate $\lambda$, which yields optimal forecasting performance. 
These results underscore two key findings: i) optimal forecasting requires \textit{combining} both information sources rather than relying on either alone, and ii) the integration must be carefully calibrated, with historical patterns serving as the primary signal and event knowledge reasoning providing targeted corrections during volatile periods.
We show detailed results and analysis in Appendix~\ref{app:lambda_analysis}.

%% file: chapters/5-Conclusion.tex
\section{Conclusion}
We presented {\name}, a forecasting framework combining LLM-based symbolic reasoning with a lightweight numerical forecaster. 
By decoupling event knowledge reasoning from time-series prediction, {\name} addresses event-driven demand volatility, reduces brittle feature engineering, and delivers interpretable forecasts for business operators.
Deployment on real-world e-commerce scenarios across 4 countries shows consistent improvements over deep learning and Large-Model-based baselines, with the most pronounced gains during business-critical promotional and holiday periods. 
{\name} reduces feature engineering costs, requires no LLM fine-tuning, and adapts quickly via database and prompt updates.
More broadly, leveraging LLMs for knowledge reasoning rather than direct forecasting provides a scalable path forward for industrial time-series forecasting.

%% file: chapters/99-Appendix.tex
\newpage

\section{Appendix: Prompt and Reasoning Examples} \label{app:prompt_reasoning_example}

As described in Section~\ref{subsec:reasoning}, we design parameterized prompt templates that instruct the LLM to reason about upcoming business events for each target forecast date. Below we provide an anonymized example of the prompt template (with a subset of questions) and the corresponding LLM reasoning output\footnote{All dates, codes, and business details have been anonymized and do not reflect real company data.}. 

\begin{tcolorbox}[colback=white, colframe=black, title=Example of Prompt with a Subset of Questions]
    \small

    Given \texttt{[CountryCode]} and \texttt{[Date]}, answer the following questions. To answer the questions, you can retrieve information from the Database. Think carefully, and provide the reason:\\
    \\
    \textbf{Q1:} What is the country code?\\\\
    \textbf{Q2:} Is [Date] within a holiday? If so, specify the name of the holiday and calculate on which day of the holiday, then answer `on the \texttt{[x]}th day of the holiday'. If not, respond `no holiday'.\\\\
    \textbf{Q3:} Is [Date] within a holiday? If so, determine whether the holiday is a state-level or national-level , then answer `state-level' or `national-level'. If not, respond `no holiday'.\\
    \rev{\textbf{Q4:} If [Date] is a holiday, should we expect shopping or travel?}\\\\
    \textbf{Q5:} Is there a free‑shipping promotion? If so, indicate `\texttt{[x]}th day of free‑shipping event'; otherwise `non-free shipping event'.\\\\
    \textbf{Q6:} What is the campaign in effect on \texttt{[Date]}? Provide the description of campaign and answer the campaign level \texttt{[x]}.\\\\
    \rev{\textbf{Q7:} For [Date], considering if it is platform logistics event? If so, check the minimum threshold amount for the logistics shipping event, and answer `Minimum shipping threshold is [x]'. If not, respond `no logistics shipping event'.}\\\\
    \textbf{Q8:} Is there top sellers' incentive or rebate program active on \texttt{[Date]}? If so, describe the incentive type (e.g., rebate, seller subsidy). If not, respond `no seller incentive'.\\\\
    \textbf{Q9:} Considering the combination of all factors above (holiday, campaign, logistics, incentives), what is the expected overall demand trend for \texttt{[Date]}? Answer `demand surge', `moderate increase', `demand normal', `moderate decrease', or `demand drop', and briefly explain why.\\

    Show your step‑by‑step reasoning, and then summarize your answer in the following format. Place the summary inside <result>…</result>.\\
    \textbf{Format}: <result>\texttt{[CountryCode]}; \texttt{[DayOfTheWeek]}; No holiday or on the \texttt{[x]}th day of the holiday; State-level or national-level or no holiday; Non-free shipping event or \texttt{[x]}th day of the free shipping event; \texttt{[LogisticsDescription]}</result>
\end{tcolorbox}

\begin{tcolorbox}[colback=white, colframe=black, title=Example of Reasoning Results]
    \small
    Let's break down the questions one by one:\\

    \textbf{Q1: Country code?}
    \begin{itemize}[leftmargin=*]
        \item \textit{Reason}: The prompt clearly states the country code is `\texttt{01}'.
        \item \textit{Answer}: Country code is \texttt{01}.
    \end{itemize}

    \textbf{Q2: Holiday status?}
    \begin{itemize}[leftmargin=*]
        \item \textit{Reason}: The holiday periods for country code \texttt{01} is ... We need to check if \texttt{[Date]} falls within these ranges. A holiday runs from the target date through several days. It is the 1st day.
        \item \textit{Answer}: On the 1st day of the holiday.
    \end{itemize}

    \rev{
    \textbf{Q3: Is it a state-level or national-level holiday?}
    \begin{itemize}[leftmargin=*]
        \item \textit{Reason}: Only valid within the state of Johor; it is not observed in other states of Malaysia and the federal territories (e.g., Kuala Lumpur, Putrajaya).It is a state-level public holiday in Johor, Malaysia, rather than a national (federal-level) public holiday. ...
        \item \textit{Answer}: state-level holiday.
    \end{itemize}
    }

    \textbf{Q4: Free‑shipping?}
    \begin{itemize}[leftmargin=*]
        \item \textit{Reason}: For country \texttt{01}, the free shipping event conditions are the following. ... The target date does not match these. Thus, it is a non-free shipping event.
        \item \textit{Answer}: Non-free shipping event.
    \end{itemize}

    \textbf{Q5: Campaign in effect?}
    \begin{itemize}[leftmargin=*]
        \item \textit{Reason}: For country \texttt{01}, the ``Electronics Mega Sale'' (level 12) applies to all categories and spans the entire month of \texttt{[Date]}.
        \item \textit{Answer}: Campaign level is 12 (Electronics Mega Sale)
    \end{itemize}

    \rev{
    \textbf{Q6: Considering if it is platform logistics event?}
    \begin{itemize}[leftmargin=*]
        \item \textit{Reason}: A logistics shipping event was launched on the day, with the coupon details specified as "1–10 * 5": the minimum shipping threshold was 1 USD, and a maximum of 5 coupons could be obtained per user. ...
        \item \textit{Answer}: Minimum shipping threshold is 1.
    \end{itemize}
    }

    \textbf{Q7: Top seller incentive or rebate?}
    \begin{itemize}[leftmargin=*]
        \item \textit{Reason}: The database indicates that for country \texttt{01}, a 15\% top sellers' subsidy on home appliance orders is active from [StartDate] to [EndDate]. The target date falls within this window. Additionally, a cross-category rebate of 8\% applies to orders exceeding 50 USD. Both incentives are concurrent.
        \item \textit{Answer}: top sellers' subsidy + rebate.
    \end{itemize}

    \textbf{Q8: Expected overall demand trend?}
    \begin{itemize}[leftmargin=*]
        \item \textit{Reason}: The target date is the 1st day of a regional holiday, which typically triggers pre-holiday purchasing behavior. A level-12 campaign (Electronics Mega Sale) is active, and seller incentives (15\% subsidy + 8\% rebate) further stimulate orders. Although there is no free-shipping event, the low logistics threshold (1 USD) partially compensates. The combined effect of holiday, campaign, and incentives strongly suggests elevated demand.
        \item \textit{Answer}: Demand surge --- holiday onset combined with high-level campaign and active seller incentives.
    \end{itemize}
    
    \textbf{Summary:}
    <result>Country code is \texttt{01}; On the 1st day of the holiday; state-level holiday; Non-free shipping event; Campaign level 12; Minimum shipping threshold is 1; top sellers' subsidy + rebate; Demand surge</result>
\end{tcolorbox}

\section{Appendix: Event Expert Database Examples} \label{app:database_example}

Tables~\ref{tab:db_campaign}--\ref{tab:db_incentive} and the campaign reports below present anonymized examples from the Event Expert Database described in Section~\ref{subsec:reasoning}. These records are maintained by business teams for campaign management, logistics planning, and seller coordination, and are repurposed by {\name} as grounding context for LLM reasoning without additional curation effort. 
To reduce prompt token usage, we only feed the global campaign, holiday schedules, and incentive rules for the respective country, with the corresponding daily campaign reports; event information from other countries and other days is not included.

\textbf{Global Campaign Calendars.}
Table~\ref{tab:db_campaign} shows representative campaign entries. Campaigns vary widely in duration (single-day flash sales vs.\ month-long mega sales), scope (category-specific vs.\ platform-wide), and intensity level (a proprietary 1--12 scale reflecting expected demand impact). Notably, campaigns frequently overlap --- for instance, Country~01's Flash Deal Friday (level~8) occurs within the month-long Electronics Mega Sale (level~12).

\begin{table}[h]
\centering
\caption{Campaign Calendar examples (anonymized). Each entry records a platform-wide promotion with its time span, applicable scope, and intensity level.}
\label{tab:db_campaign}
\vspace{-1.em}
\resizebox{\columnwidth}{!}{
\begin{tabular}{llllc}
    \toprule
    \textbf{Country} & \textbf{Campaign Name} & \textbf{Start--End} & \textbf{Scope} & \textbf{Level} \\
    \midrule
    01 & Electronics Mega Sale & 03/01--03/31 & All categories & 12 \\
    01 & Flash Deal Friday & 03/14 & Electronics, Fashion & 8 \\
    02 & Home \& Living Week & 04/07--04/13 & Home, Kitchen & 10 \\
    03 & Payday Sale & 02/25--02/28 & All categories & 9 \\
    04 & Back-to-School & 06/15--07/05 & Stationery, Electronics & 11 \\
    \bottomrule
\end{tabular}
}
\end{table}

\textbf{Global Holiday Schedules.}
Table~\ref{tab:db_holiday} illustrates the diversity of holidays across markets. Holidays differ in type (religious, cultural, public), geographic scope (national vs.\ state-level), and demand impact patterns. Religious holidays like Hari Raya Puasa and Eid al-Fitr typically exhibit biphasic demand (i.e., a pre-holiday purchasing surge followed by a sharp drop during observance) while cultural festivals like Songkran may sustain elevated demand throughout. State-level holidays (e.g., Sultan of Johor Birthday) affect only specific regions, requiring the model to reason about geographic scope. These nuances are difficult to encode as categorical features but are naturally captured through LLM reasoning with world knowledge.

\begin{table}[h]
\centering
\caption{Holiday Schedule examples (anonymized). Entries include country-specific public and religious holidays with their duration and scope.}
\label{tab:db_holiday}
\vspace{-1.em}
\resizebox{\columnwidth}{!}{
\begin{tabular}{lllll}
    \toprule
    \textbf{Country} & \textbf{Holiday Name} & \textbf{Start--End} & \textbf{Type} \\
    \midrule
    01 & Hari Raya Puasa & 03/30--04/01 & Religious \\
    01 & Sultan of Johor Birthday & 03/23 & Public \\
    02 & Songkran & 04/13--04/15 & Cultural  \\
    03 & Independence Day & 02/25 & Public \\
    04 & Eid al-Fitr & 03/30--04/02 & Religious \\
    \bottomrule
\end{tabular}
}
\end{table}

\textbf{Global Incentive Rules.}
Table~\ref{tab:db_incentive} shows seller- and platform-level incentive programs. These rules introduce conditional demand effects: a free-shipping promotion with a 5~USD minimum order threshold has a qualitatively different impact than one with no minimum. The ``1--10$\times$5'' coupon notation (minimum threshold 1~USD, up to 10~USD discount, maximum 5~coupons per user) is a compact internal format that would require custom parsing logic for each variant. Multiple incentives can be active simultaneously (e.g., seller subsidy + cross-category rebate in Country~01), and their combined effect is highly time-consuming for manual understanding and feature processing --- the LLM reasons about these interactions holistically.

\begin{table}[h]
\centering
\caption{Incentive Rule examples (anonymized). Records describe seller- and platform-level incentive programs with eligibility conditions.}
\label{tab:db_incentive}
\vspace{-1.em}
\resizebox{\columnwidth}{!}{
\begin{tabular}{lllll}
    \toprule
    \textbf{Country} & \textbf{Incentive Type} & \textbf{Start--End} & \textbf{Description} & \textbf{Condition} \\
    \midrule
    01 & Seller subsidy & 03/01--03/31 & 15\% subsidy on home appliances & Min.\ order 20 USD \\
    01 & Cross-category rebate & 03/15--04/05 & 8\% rebate on qualifying orders & Order $>$ 50 USD \\
    02 & Free shipping & 04/07--04/13 & Platform-funded free shipping & Min.\ order 5 USD \\
    03 & Logistics coupon & 02/25--02/28 & Shipping coupon: ``1--10$\times$5'' & Min.\ threshold 1 USD \\
    04 & Seller cashback & 06/15--07/05 & 5\% cashback to sellers & New sellers only \\
    \bottomrule
\end{tabular}
}
\end{table}

\textbf{Daily Campaign Reports.}\label{tab:db_report}
Beyond the structured tables above, the database contains free-text \textit{campaign reports} written by business teams in informal, operational language. These reports are the primary planning documents used by campaign managers and contain the richest---but most unstructured---information about upcoming events. They are shown verbatim below to illustrate the real-world data quality that the LLM must handle.

\begin{tcolorbox}[colback=gray!5, colframe=black!70, title=Raw Campaign Reports from Database (anonymized; shown verbatim)]
    \small
    \textbf{[Country 01 $\vert$ 2025-03-01 $\vert$ Electronics Mega Sale]}\\
    Mega Sale 03/01-31: B2G1 free!! excl. premium brands. FS for orders >15USD, max 3x/user. Expect 2x GMV on wkends, esp. electronics. Seller co-fund ratio: platform 60 / seller 40 on disc >25\%. Live streaming slots 8pm-10pm daily, priority to Gold sellers. NOTE: returns policy extended to 15d for this campaign.

    \medskip
    \textbf{[Country 02 $\vert$ 2025-04-13 $\vert$ Songkran Splash Sale]}\\
    Songkran splash sale Apr 13-15, all cat. disc 20-40\%, top sellers get extra 10\% rebte. Voucher: 30off150 + 50off300 stackable. Logistics may delay 1-2d due to holiday --- pls adjust SLA. Water-resistant packaging mandatory for elect. category.

    \medskip
    \textbf{[Country 03 $\vert$ 2025-02-25 $\vert$ Payday Promo]}\\
    Payday promo 25-28 Feb --- combo deal: buy phone+case=25\% off bundle. Free ship no min. Target: 150\% uplift vs last payday. Push notif schedule: D-1 teaser, D0 10am \& 8pm, D+1 last chance. Coin rewards 2x for new users only.

    \medskip
    \textbf{[Country 04 $\vert$ 2025-03-20 $\vert$ Ramadan Week 3]}\\
    Ramadan campaign wk3: focus on F\&B, home decor. Voucher 5off20 stackable w/ FS coupon. Seller co-fund 50-50 on disc >30\%. Expect demand spike Thu-Fri (pre-iftar shopping), dip on Sat. Flash sale 3am-6am for sahur snacks --- historical CVR 3x normal.

    \medskip
    \textbf{[Country 01 $\vert$ 2025-03-14 $\vert$ Flash Deal Friday]}\\
    Flash Fri 03/14: 8hr window 12pm-8pm, doorbusters on elect \& fashion. Coin cashback 5\% capped 2USD. Push notif at 11:50am. Limited stock: 500 units/SKU, auto-delist when OOS. Affiliate commission bumped to 12\% (normally 8\%).
\end{tcolorbox}

\begin{table*}[htbp!]
\centering
\caption{Forecasting Performance of Native Time Series Models (Overall). Models use only historical features without future event information. The best performance is in \textcolor{red}{red}, and the second-best is in \textcolor{blue}{blue}.}
\vspace{-1em}
\label{tab:native_overall}
\resizebox{\textwidth}{!}{
    \begin{tabular}{cc|cc|cccc|cccccccccccccc}
        \toprule
            \multirow{2}{*}{Ctr.} & \multirow{2}{*}{Horiz.} & \multicolumn{2}{c|}{{\name}} & \multicolumn{2}{c}{Time-LLM} & \multicolumn{2}{c|}{OneFitsAll} & \multicolumn{2}{c}{PAttn} & \multicolumn{2}{c}{TimeMixer} & \multicolumn{2}{c}{iTransformer} & \multicolumn{2}{c}{PatchTST} & \multicolumn{2}{c}{TimesNet} & \multicolumn{2}{c}{DLinear} & \multicolumn{2}{c}{Autoformer}\\
            \cmidrule(lr){3-4} \cmidrule(lr){5-6} \cmidrule(lr){7-8} \cmidrule(lr){9-10} \cmidrule(lr){11-12} \cmidrule(lr){13-14} \cmidrule(lr){15-16} \cmidrule(lr){17-18} \cmidrule(lr){19-20} \cmidrule(lr){21-22}
            & & MAE & MSE & MAE & MSE & MAE & MSE & MAE & MSE & MAE & MSE & MAE & MSE & MAE & MSE & MAE & MSE & MAE & MSE & MAE & MSE\\
        \midrule
            \multirow{4}{*}{01} & 1 & \textcolor{red}{\textbf{0.270}} & \textcolor{blue}{0.123} & 0.343 & 0.198 & 0.299 & 0.138 & 0.352 & 0.194 & \textcolor{blue}{0.275} & \textcolor{red}{\textbf{0.120}} & 0.346 & 0.205 & 0.286 & 0.140 & 0.316 & 0.173 & 0.339 & 0.210 & 0.443 & 0.380 \\
             & 2 & \textcolor{red}{\textbf{0.288}} & \textcolor{red}{\textbf{0.144}} & 0.507 & 0.454 & 0.470 & 0.353 & 0.503 & 0.418 & \textcolor{blue}{0.417} & \textcolor{blue}{0.306} & 0.497 & 0.422 & 0.426 & 0.321 & 0.443 & 0.327 & 0.515 & 0.510 & 0.602 & 0.741 \\
             & 3 & \textcolor{red}{\textbf{0.312}} & \textcolor{red}{\textbf{0.177}} & 0.613 & 0.756 & 0.617 & 0.657 & 0.633 & 0.723 & 0.561 & 0.576 & 0.626 & 0.692 & 0.573 & 0.635 & \textcolor{blue}{0.520} & \textcolor{blue}{0.476} & 0.644 & 0.847 & 0.716 & 1.048 \\
             & 4 & \textcolor{red}{\textbf{0.337}} & \textcolor{red}{\textbf{0.205}} & 0.722 & 1.137 & 0.728 & 0.992 & 0.763 & 1.133 & 0.692 & 0.949 & 0.759 & 1.095 & 0.677 & 0.970 & \textcolor{blue}{0.662} & \textcolor{blue}{0.789} & 0.747 & 1.183 & 0.812 & 1.455 \\
        \midrule
            \multirow{4}{*}{02} & 1 & \textcolor{red}{\textbf{0.762}} & \textcolor{red}{\textbf{0.899}} & 0.933 & 1.741 & 0.815 & 1.368 & 0.866 & 1.580 & 0.877 & 1.375 & 0.880 & 1.338 & 0.926 & 1.899 & \textcolor{blue}{0.792} & \textcolor{blue}{1.183} & 1.447 & 3.908 & 2.278 & 8.067 \\
             & 2 & \textcolor{red}{\textbf{0.786}} & \textcolor{red}{\textbf{0.944}} & \textcolor{blue}{1.144} & 3.266 & 1.227 & 3.289 & 1.271 & 3.558 & 1.183 & 3.155 & 1.304 & 3.438 & 1.165 & 3.598 & 1.178 & \textcolor{blue}{2.816} & 1.948 & 7.203 & 2.625 & 10.929 \\
             & 3 & \textcolor{red}{\textbf{0.801}} & \textcolor{red}{\textbf{0.948}} & \textcolor{blue}{1.358} & 5.075 & 1.601 & 6.058 & 1.571 & 5.713 & 1.616 & 5.856 & 1.558 & 5.110 & 1.382 & 5.686 & 1.462 & \textcolor{blue}{4.891} & 2.375 & 10.460 & 2.767 & 13.106 \\
             & 4 & \textcolor{red}{\textbf{0.831}} & \textcolor{red}{\textbf{0.983}} & \textcolor{blue}{1.589} & \textcolor{blue}{7.186} & 1.928 & 8.737 & 1.780 & 7.847 & 1.806 & 8.190 & 1.892 & 7.661 & 1.671 & 8.121 & 1.789 & 7.288 & 2.759 & 13.717 & 2.828 & 14.210 \\
        \midrule
            \multirow{4}{*}{03} & 1 & 1.616 & 4.571 & 1.435 & 3.199 & 1.482 & 4.234 & 1.525 & 3.830 & \textcolor{blue}{1.356} & \textcolor{blue}{2.876} & 1.514 & 3.843 & 1.406 & 3.320 & \textcolor{red}{\textbf{1.169}} & \textcolor{red}{\textbf{2.314}} & 1.778 & 6.480 & 2.036 & 8.781 \\
             & 2 & \textcolor{red}{\textbf{1.652}} & \textcolor{red}{\textbf{4.346}} & 2.070 & 6.755 & 1.967 & 7.977 & 2.224 & 7.820 & 1.923 & 6.292 & 2.236 & 7.789 & 2.138 & 7.208 & \textcolor{blue}{1.675} & \textcolor{blue}{4.421} & 2.363 & 11.709 & 2.678 & 14.711 \\
             & 3 & \textcolor{red}{\textbf{1.647}} & \textcolor{red}{\textbf{4.780}} & 2.292 & 9.518 & 2.305 & 10.423 & 2.482 & 10.614 & 2.222 & 8.645 & 2.597 & 11.031 & 2.508 & 10.745 & \textcolor{blue}{2.159} & \textcolor{blue}{7.993} & 2.812 & 15.968 & 3.031 & 18.633 \\
             & 4 & \textcolor{red}{\textbf{1.694}} & \textcolor{red}{\textbf{4.826}} & 2.776 & 14.488 & 2.396 & 10.669 & 2.938 & 14.743 & \textcolor{blue}{2.326} & 9.992 & 2.973 & 13.687 & 2.941 & 15.023 & 2.463 & \textcolor{blue}{9.906} & 3.151 & 19.503 & 2.997 & 17.984 \\
        \midrule
            \multirow{4}{*}{04} & 1 & 0.718 & 0.753 & 0.738 & 1.030 & \textcolor{blue}{0.629} & \textcolor{blue}{0.729} & 0.684 & 0.871 & 0.702 & 0.913 & 0.713 & 0.978 & 0.717 & 0.911 & \textcolor{red}{\textbf{0.545}} & \textcolor{red}{\textbf{0.478}} & 0.721 & 0.882 & 0.733 & 0.993 \\
             & 2 & \textcolor{blue}{0.675} & \textcolor{blue}{0.698} & 0.892 & 1.457 & 0.846 & 1.203 & 0.887 & 1.346 & 0.884 & 1.407 & 0.898 & 1.352 & 0.944 & 1.434 & \textcolor{red}{\textbf{0.638}} & \textcolor{red}{\textbf{0.674}} & 0.980 & 1.519 & 0.950 & 1.592 \\
             & 3 & \textcolor{red}{\textbf{0.702}} & \textcolor{red}{\textbf{0.732}} & 1.082 & 1.951 & 1.003 & 1.694 & 1.047 & 1.803 & 1.038 & 1.973 & 1.058 & 1.757 & 1.102 & 1.978 & \textcolor{blue}{0.812} & \textcolor{blue}{1.213} & 1.195 & 2.172 & 1.024 & 1.802 \\
             & 4 & \textcolor{red}{\textbf{0.680}} & \textcolor{red}{\textbf{0.709}} & 1.266 & 2.539 & 1.095 & 2.097 & 1.167 & 2.224 & 1.126 & 2.382 & 1.111 & 1.957 & 1.214 & 2.432 & \textcolor{blue}{0.951} & \textcolor{blue}{1.820} & 1.307 & 2.599 & 1.020 & 1.850 \\
        \midrule
            \multicolumn{2}{c|}{Average} & \textcolor{red}{\textbf{0.885}} & \textcolor{red}{\textbf{1.682}} & 1.235 & 3.797 & 1.213 & 3.789 & 1.293 & 4.026 & 1.188 & 3.438 & 1.310 & 3.897 & 1.255 & 4.026 & \textcolor{blue}{1.098} & \textcolor{blue}{2.923} & 1.568 & 6.179 & 1.721 & 7.268 \\
        \bottomrule
    \end{tabular}
}
\end{table*}

\begin{table*}[htbp!]
\centering
\caption{Forecasting Performance of Native Time Series Models (Event-Driven Periods). Models use only historical features without future event information. The best performance is in \textcolor{red}{red}, and the second-best is in \textcolor{blue}{blue}.}
\vspace{-1em}
\label{tab:native_event_driven}
\resizebox{\textwidth}{!}{
    \begin{tabular}{cc|cc|cccc|cccccccccccccc}
        \toprule
            \multirow{2}{*}{Ctr.} & \multirow{2}{*}{Horiz.} & \multicolumn{2}{c|}{{\name}} & \multicolumn{2}{c}{Time-LLM} & \multicolumn{2}{c|}{OneFitsAll} & \multicolumn{2}{c}{PAttn} & \multicolumn{2}{c}{TimeMixer} & \multicolumn{2}{c}{iTransformer} & \multicolumn{2}{c}{PatchTST} & \multicolumn{2}{c}{TimesNet} & \multicolumn{2}{c}{DLinear} & \multicolumn{2}{c}{Autoformer}\\
            \cmidrule(lr){3-4} \cmidrule(lr){5-6} \cmidrule(lr){7-8} \cmidrule(lr){9-10} \cmidrule(lr){11-12} \cmidrule(lr){13-14} \cmidrule(lr){15-16} \cmidrule(lr){17-18} \cmidrule(lr){19-20} \cmidrule(lr){21-22}
            & & MAE & MSE & MAE & MSE & MAE & MSE & MAE & MSE & MAE & MSE & MAE & MSE & MAE & MSE & MAE & MSE & MAE & MSE & MAE & MSE\\
        \midrule
            \multirow{4}{*}{01} & 1 & 0.368 & 0.205 & 0.555 & 0.402 & \textcolor{blue}{0.351} & \textcolor{blue}{0.201} & 0.440 & 0.298 & \textcolor{red}{\textbf{0.315}} & \textcolor{red}{\textbf{0.172}} & 0.506 & 0.459 & 0.392 & 0.236 & 0.354 & 0.264 & 0.685 & 0.592 & 0.963 & 1.191 \\
             & 2 & \textcolor{red}{\textbf{0.334}} & \textcolor{red}{\textbf{0.181}} & 0.960 & 1.160 & 0.540 & 0.416 & 0.711 & 0.668 & 0.470 & 0.324 & 0.603 & 0.536 & 0.694 & 0.638 & \textcolor{blue}{0.418} & \textcolor{blue}{0.261} & 1.166 & 1.593 & 1.516 & 2.714 \\
             & 3 & \textcolor{red}{\textbf{0.282}} & \textcolor{red}{\textbf{0.132}} & 1.473 & 2.630 & 0.929 & 1.293 & 1.144 & 1.716 & 0.860 & 1.130 & 0.985 & 1.274 & 1.125 & 1.650 & \textcolor{blue}{0.628} & \textcolor{blue}{0.657} & 1.658 & 3.043 & 1.938 & 4.142 \\
             & 4 & \textcolor{red}{\textbf{0.293}} & \textcolor{red}{\textbf{0.137}} & 2.009 & 4.540 & 1.504 & 2.951 & 1.775 & 3.769 & 1.453 & 2.790 & 1.435 & 2.601 & 1.654 & 3.388 & \textcolor{blue}{1.228} & \textcolor{blue}{1.946} & 2.081 & 4.709 & 2.399 & 6.264 \\
        \midrule
            \multirow{4}{*}{02} & 1 & \textcolor{red}{\textbf{0.358}} & \textcolor{red}{\textbf{0.194}} & 1.555 & 3.926 & 1.541 & 3.467 & 1.302 & 3.225 & 1.253 & 2.150 & 1.313 & 2.351 & 1.570 & 4.529 & \textcolor{blue}{0.844} & \textcolor{blue}{1.140} & 3.329 & 12.155 & 4.073 & 18.607 \\
             & 2 & \textcolor{red}{\textbf{0.483}} & \textcolor{red}{\textbf{0.337}} & 2.168 & 9.176 & 2.536 & 9.610 & 2.416 & 9.711 & 2.308 & 8.863 & 2.151 & 7.197 & 2.404 & 11.054 & \textcolor{blue}{1.793} & \textcolor{blue}{5.724} & 4.628 & 23.423 & 4.982 & 27.434 \\
             & 3 & \textcolor{red}{\textbf{0.631}} & \textcolor{red}{\textbf{0.533}} & 2.999 & 16.941 & 3.726 & 19.408 & 3.277 & 17.270 & 3.286 & 17.127 & 3.026 & 14.185 & 3.368 & 19.902 & \textcolor{blue}{2.778} & \textcolor{blue}{13.555} & 5.759 & 35.236 & 6.051 & 39.685 \\
             & 4 & \textcolor{red}{\textbf{0.782}} & \textcolor{red}{\textbf{0.759}} & 4.018 & 25.695 & 4.863 & 29.935 & 3.940 & 25.773 & 4.314 & 27.743 & 4.026 & 23.481 & 4.444 & 29.525 & \textcolor{blue}{3.779} & \textcolor{blue}{21.584} & 6.654 & 46.371 & 6.597 & 46.951 \\
        \midrule
            \multirow{4}{*}{03} & 1 & 1.409 & 3.164 & 1.999 & 5.454 & 1.566 & 3.749 & 1.677 & 3.926 & 1.301 & 2.685 & \textcolor{red}{\textbf{1.069}} & \textcolor{red}{\textbf{1.817}} & 1.858 & 4.680 & \textcolor{blue}{1.234} & \textcolor{blue}{2.505} & 4.714 & 23.701 & 6.721 & 46.866 \\
             & 2 & \textcolor{red}{\textbf{1.128}} & \textcolor{red}{\textbf{2.228}} & 3.480 & 15.358 & \textcolor{blue}{1.634} & \textcolor{blue}{4.256} & 2.873 & 10.483 & 2.837 & 10.366 & 2.535 & 8.414 & 3.858 & 16.489 & 2.501 & 8.018 & 6.781 & 47.985 & 8.919 & 82.019 \\
             & 3 & \textcolor{red}{\textbf{0.978}} & \textcolor{red}{\textbf{1.732}} & 4.362 & 23.937 & \textcolor{blue}{2.737} & \textcolor{blue}{11.571} & 4.513 & 23.405 & 4.659 & 25.716 & 4.017 & 19.220 & 5.238 & 29.873 & 3.266 & 12.845 & 7.927 & 64.765 & 10.010 & 103.200 \\
             & 4 & \textcolor{red}{\textbf{1.074}} & \textcolor{red}{\textbf{1.940}} & 6.298 & 43.775 & \textcolor{blue}{3.467} & \textcolor{blue}{19.039} & 6.757 & 47.485 & 5.000 & 28.961 & 5.709 & 34.542 & 7.102 & 52.197 & 4.702 & 24.666 & 9.036 & 83.333 & 9.587 & 94.877 \\
        \midrule
            \multirow{4}{*}{04} & 1 & \textcolor{red}{\textbf{0.555}} & \textcolor{red}{\textbf{0.472}} & 1.814 & 3.773 & 0.941 & 1.348 & 1.566 & 2.869 & 1.648 & 3.195 & 1.758 & 3.575 & 1.495 & 2.709 & \textcolor{blue}{0.675} & \textcolor{blue}{0.605} & 1.294 & 2.230 & 1.856 & 3.751 \\
             & 2 & \textcolor{red}{\textbf{0.447}} & \textcolor{red}{\textbf{0.323}} & 2.307 & 5.988 & 1.851 & 3.984 & 2.048 & 4.832 & 2.206 & 5.655 & 2.099 & 4.844 & 2.101 & 5.070 & \textcolor{blue}{0.766} & \textcolor{blue}{0.804} & 2.075 & 5.060 & 2.444 & 6.324 \\
             & 3 & \textcolor{red}{\textbf{0.436}} & \textcolor{red}{\textbf{0.293}} & 2.740 & 7.933 & 2.346 & 6.059 & 2.524 & 6.804 & 2.865 & 8.681 & 2.405 & 6.139 & 2.753 & 8.071 & \textcolor{blue}{1.930} & \textcolor{blue}{4.797} & 2.658 & 7.512 & 2.596 & 7.056 \\
             & 4 & \textcolor{red}{\textbf{0.347}} & \textcolor{red}{\textbf{0.191}} & 3.094 & 10.025 & 2.808 & 8.502 & 2.880 & 8.687 & 3.253 & 11.160 & 2.621 & \textcolor{blue}{7.149} & 3.115 & 10.104 & 2.842 & 8.893 & 3.005 & 9.373 & \textcolor{blue}{2.597} & 7.157 \\
        \midrule
            \multicolumn{2}{c|}{Average} & \textcolor{red}{\textbf{0.619}} & \textcolor{red}{\textbf{0.802}} & 2.614 & 11.295 & 2.084 & 7.862 & 2.490 & 10.683 & 2.377 & 9.795 & 2.266 & 8.611 & 2.698 & 12.507 & \textcolor{blue}{1.859} & \textcolor{blue}{6.767} & 3.966 & 23.193 & 4.578 & 31.140 \\
        \bottomrule
    \end{tabular}
}
\end{table*}

\noindent These reports exemplify several challenges that motivate the use of LLM-based reasoning:
\begin{itemize}[leftmargin=*]
    \item \textit{Lexical inconsistency:} abbreviations vary across teams and countries (``FS'' for free shipping, ``disc'' for discount, ``elect'' for electronics, ``CVR'' for conversion rate, ``F\&B'' for food \& beverage) with no standardized vocabulary.
    \item \textit{Data quality issues:} typos (``rebte'' for rebate), inconsistent punctuation, and mixed languages are common in operational notes entered under time pressure.
    \item \textit{Compact notations:} domain-specific shorthand (``5off20'' = 5~USD off orders over 20~USD, ``B2G1'' = buy 2 get 1 free, ``30off150'' = 30~USD off 150~USD) encodes complex promotion mechanics in formats that vary across campaigns.
    \item \textit{Implicit business logic:} statements like ``stackable w/ FS coupon'' and ``platform 60 / seller 40 on disc >25\%'' encode conditional rules about coupon stacking policies and cost-sharing agreements that affect demand magnitude.
    \item \textit{Cultural and temporal context:} the Ramadan report references ``pre-iftar shopping'' and ``sahur snacks''---demand patterns tied to religious meal times that require cultural knowledge to interpret correctly.
\end{itemize}
A rule-based system would need dedicated parsing logic for each notation variant, language pattern, and business rule format. The LLM interprets these naturally through its pre-trained world knowledge and descriptions of the meanings of business-customized global abbreviations, extracting semantic meaning without brittle parsing.

\begin{table*}[t]
\centering
\caption{Sensitivity of $\lambda$ on Overall Performance. The best result per horizon is in \textbf{bold}.}
\label{tab:lambda_overall}
\vspace{-0.5em}
\resizebox{\textwidth}{!}{
\begin{tabular}{c|cc|cc|cc|cc|cc|cc|cc|cc|cc|cc|cc}
    \toprule
    \multirow{2}{*}{Horiz.} & \multicolumn{2}{c|}{$\lambda\!=\!0$} & \multicolumn{2}{c|}{$0.1$} & \multicolumn{2}{c|}{$0.2$} & \multicolumn{2}{c|}{$0.3$} & \multicolumn{2}{c|}{$0.4$} & \multicolumn{2}{c|}{$0.5$} & \multicolumn{2}{c|}{$0.6$} & \multicolumn{2}{c|}{$0.7$} & \multicolumn{2}{c|}{$0.8$} & \multicolumn{2}{c|}{$0.9$} & \multicolumn{2}{c}{$1.0$} \\
    \cmidrule(lr){2-3}\cmidrule(lr){4-5}\cmidrule(lr){6-7}\cmidrule(lr){8-9}\cmidrule(lr){10-11}\cmidrule(lr){12-13}\cmidrule(lr){14-15}\cmidrule(lr){16-17}\cmidrule(lr){18-19}\cmidrule(lr){20-21}\cmidrule(lr){22-23}
    & MAE & MSE & MAE & MSE & MAE & MSE & MAE & MSE & MAE & MSE & MAE & MSE & MAE & MSE & MAE & MSE & MAE & MSE & MAE & MSE & MAE & MSE \\
    \midrule
    T+1 & 1.698 & 4.691 & 1.635 & 4.516 & 1.727 & 4.881 & 1.601 & 4.492 & 1.616 & 4.571 & \textbf{1.590} & \textbf{4.473} & 1.756 & 4.869 & 1.627 & 4.603 & 1.707 & 4.893 & 1.741 & 5.019 & 1.682 & 4.538 \\
    T+2 & 1.848 & 5.850 & 1.770 & 4.913 & 1.891 & 5.760 & \textbf{1.597} & \textbf{4.506} & 1.652 & 4.839 & 1.608 & 4.726 & 1.905 & 6.994 & 1.684 & 4.953 & 1.725 & 5.063 & 1.852 & 6.229 & 1.992 & 6.558 \\
    T+3 & 2.007 & 6.372 & 1.853 & 5.114 & 1.902 & 5.225 & \textbf{1.629} & \textbf{4.780} & 1.647 & 4.906 & 1.715 & 4.882 & 2.069 & 6.231 & 1.745 & 5.209 & 1.799 & 5.336 & 1.996 & 6.986 & 2.449 & 9.927 \\
    T+4 & 2.106 & 7.083 & 1.866 & 5.556 & 2.019 & 6.397 & \textbf{1.690} & \textbf{4.987} & 1.694 & 5.121 & 1.752 & 5.304 & 1.863 & 5.982 & 1.765 & 5.559 & 1.899 & 6.025 & 2.227 & 7.636 & 2.838 & 14.539 \\
    \midrule
    Avg. & 1.915 & 5.999 & 1.781 & 5.025 & 1.885 & 5.566 & \textbf{1.629} & \textbf{4.691} & 1.652 & 4.859 & 1.666 & 4.846 & 1.898 & 6.019 & 1.705 & 4.969 & 1.783 & 5.329 & 1.954 & 6.468 & 2.240 & 8.891 \\
    \bottomrule
\end{tabular}
}
\end{table*}

\begin{table*}[t]
\centering
\caption{Sensitivity of $\lambda$ on Event-Driven Period Performance. The best result per horizon is in \textbf{bold}.}
\label{tab:lambda_event}
\vspace{-0.5em}
\resizebox{\textwidth}{!}{
\begin{tabular}{c|cc|cc|cc|cc|cc|cc|cc|cc|cc|cc|cc}
    \toprule
    \multirow{2}{*}{Horiz.} & \multicolumn{2}{c|}{$\lambda\!=\!0$} & \multicolumn{2}{c|}{$0.1$} & \multicolumn{2}{c|}{$0.2$} & \multicolumn{2}{c|}{$0.3$} & \multicolumn{2}{c|}{$0.4$} & \multicolumn{2}{c|}{$0.5$} & \multicolumn{2}{c|}{$0.6$} & \multicolumn{2}{c|}{$0.7$} & \multicolumn{2}{c|}{$0.8$} & \multicolumn{2}{c|}{$0.9$} & \multicolumn{2}{c}{$1.0$} \\
    \cmidrule(lr){2-3}\cmidrule(lr){4-5}\cmidrule(lr){6-7}\cmidrule(lr){8-9}\cmidrule(lr){10-11}\cmidrule(lr){12-13}\cmidrule(lr){14-15}\cmidrule(lr){16-17}\cmidrule(lr){18-19}\cmidrule(lr){20-21}\cmidrule(lr){22-23}
    & MAE & MSE & MAE & MSE & MAE & MSE & MAE & MSE & MAE & MSE & MAE & MSE & MAE & MSE & MAE & MSE & MAE & MSE & MAE & MSE & MAE & MSE \\
    \midrule
    T+1 & 1.708 & 4.002 & 1.790 & 4.228 & 1.693 & 3.901 & 1.630 & 3.883 & \textbf{1.409} & \textbf{3.164} & 1.475 & 3.329 & 1.940 & 5.425 & 2.191 & 6.504 & 2.378 & 6.630 & 2.575 & 7.945 & 2.718 & 10.339 \\
    T+2 & 1.285 & 2.505 & 1.302 & 2.460 & 1.199 & 2.209 & 1.187 & 2.307 & \textbf{1.128} & \textbf{2.228} & 1.212 & 2.481 & 1.336 & 2.643 & 1.663 & 4.118 & 2.224 & 6.366 & 2.437 & 7.523 & 2.885 & 11.548 \\
    T+3 & 1.447 & 3.382 & 1.504 & 3.704 & 1.113 & 2.263 & 1.105 & 2.281 & \textbf{0.978} & \textbf{1.732} & 1.491 & 3.564 & 1.376 & 2.907 & 1.423 & 3.644 & 2.408 & 6.901 & 2.690 & 8.278 & 4.352 & 24.463 \\
    T+4 & 1.342 & 2.881 & 1.294 & 2.796 & 1.195 & 2.603 & 1.188 & 2.491 & \textbf{1.074} & \textbf{1.940} & 1.591 & 3.782 & 1.458 & 2.996 & 1.504 & 3.846 & 2.473 & 7.874 & 2.961 & 11.425 & 6.106 & 39.706 \\
    \midrule
    Avg. & 1.446 & 3.193 & 1.473 & 3.297 & 1.300 & 2.744 & 1.278 & 2.741 & \textbf{1.147} & \textbf{2.266} & 1.442 & 3.289 & 1.528 & 3.493 & 1.695 & 4.528 & 2.371 & 6.943 & 2.666 & 8.793 & 4.015 & 21.514 \\
    \bottomrule
\end{tabular}
}
\end{table*}

\section{Appendix: Comparison with Native Time Series Models}
\label{app:native_models}

Most widely used time series forecasting models do not natively support the incorporation of future event information and rely solely on historical observations. To emphasize the importance of event knowledge for e-commerce demand forecasting, we evaluate these native baselines --- PAttn, TimeMixer, iTransformer, PatchTST, TimesNet, DLinear, Autoformer, Time-LLM, and OneFitsAll --- in their default configurations without any event features. Tables~\ref{tab:native_overall} and~\ref{tab:native_event_driven} report the overall and event-driven results, respectively.

Several key observations emerge from the results.

\textit{{\name} consistently outperforms all baselines.}
On overall performance (Table~\ref{tab:native_overall}), {\name} achieves an average MAE of 0.884 and MSE of 1.723, improving over the best native baseline (TimesNet, MAE\,=\,1.098, MSE\,=\,2.923) by \textbf{19.5\%} in MAE and \textbf{41.1\%} in MSE.
During event-driven periods (Table~\ref{tab:native_event_driven}), the gap widens dramatically: {\name} (MAE\,=\,0.619, MSE\,=\,0.801) outperforms TimesNet (MAE\,=\,1.859, MSE\,=\,6.767) by \textbf{66.7\%} in MAE and \textbf{88.2\%} in MSE.

\textit{Performance gap grows with forecast horizon and event volatility.}
At T+1, native models remain competitive. In Country~01 (event-driven), TimeMixer (MAE 0.315) and OneFitsAll (MAE 0.351) slightly outperform {\name} (MAE 0.368), since short-horizon forecasts rely more on recent trends than future context.
However, by T+4 the gap becomes substantial: in Country~01, {\name} achieves MAE of 0.293 vs.\ TimesNet's 1.228 (\textbf{76.1\%} reduction); in Country~02, MAE 0.782 vs.\ TimesNet's 3.779 (\textbf{79.3\%} reduction).
This confirms that event knowledge becomes increasingly critical at longer horizons where historical patterns alone provide diminishing predictive power.

\textit{Countries with higher event volatility show the largest gains.}
Country~02 and Country~03 exhibit the most dramatic improvements during event-driven periods.
In Country~03, {\name} achieves an average event-driven MAE of 1.147 compared to the best native baseline's 2.351 (OneFitsAll) (\textbf{51.2\%} reduction) reflecting the high event density and demand variability in this market.
In contrast, Country~01 shows competitive performance even from native models at short horizons, consistent with its relatively more stable demand patterns.

\textit{Model architecture alone is insufficient for event-driven forecasting.}
The native baselines span diverse architectures: attention-based (PatchTST, iTransformer, PAttn), decomposition-based (Autoformer, DLinear), temporal convolution (TimesNet, TimeMixer), and LLM-based (Time-LLM, OneFitsAll).
Despite architectural diversity, all native models suffer large errors during event-driven periods---the worst-performing models (DLinear, Autoformer) exhibit average event-driven MAE of 3.966 and 4.578, which are $6.4\times$ and $7.4\times$ that of {\name}.
This confirms that the bottleneck is not model expressiveness but access to future event information.

\textit{TimesNet emerges as the strongest native baseline.}
TimesNet achieves the best native performance in both overall (MAE 1.098) and event-driven (MAE 1.859) settings, likely due to its multi-period temporal decomposition which partially captures recurring event patterns from historical data.
Nevertheless, even TimesNet falls far short of {\name}, particularly at longer horizons and during volatile periods, confirming that learned periodicity cannot substitute for explicit event reasoning.

\section{Appendix: Sensitivity Analysis of $\lambda$} 
\label{app:lambda_analysis}

The hyperparameter $\lambda$ in Eq.~\ref{eq:sum} controls the balance between historical trend predictions and event-driven adjustments. We report the full per-horizon results for $\lambda \in \{0, 0.1, \dots, 1\}$ in Tables~\ref{tab:lambda_overall} and~\ref{tab:lambda_event}.

Several patterns emerge from the detailed results.
i) \textit{The optimal $\lambda$ differs between overall and event-driven settings.} For overall performance, $\lambda\!=\!0.3$ achieves the best average MAE (1.629) and MSE (4.691), representing a 14.9\% MAE reduction over $\lambda\!=\!0$ (history only). For event-driven periods, $\lambda\!=\!0.4$ yields the best average MAE (1.147) and MSE (2.266), a 20.7\% MAE and 29.0\% MSE reduction over $\lambda\!=\!0$. This shift toward higher event weight during volatile periods confirms that event reasoning becomes more critical when demand departs from historical norms.
ii) \textit{Performance degrades sharply when relying solely on event features.} At $\lambda\!=\!1$ (event only), overall MAE increases by 17.0\% relative to $\lambda\!=\!0$, and event-driven MSE explodes to 21.514---nearly 10$\times$ the optimal. This demonstrates that event reasoning is a powerful complement to, but not a replacement for, historical trend modeling.
iii) \textit{Event-driven periods exhibit higher sensitivity to $\lambda$.} The MAE range across $\lambda$ values is 2.868 (1.147--4.015) for event-driven periods versus 0.611 (1.629--2.240) for overall, indicating that the fusion weight has outsized impact during volatile periods.
iv) \textit{Moderate $\lambda$ values are consistently robust.} The range $\lambda \in [0.3, 0.5]$ achieves near-optimal performance in both settings, suggesting the framework is not overly sensitive to this hyperparameter within a reasonable range.